\documentclass[a4paper,fleqn]{cas-sc}
\usepackage{etoolbox}
\usepackage[authoryear]{natbib}
\usepackage{caption}
\usepackage{subcaption}
\usepackage{stfloats}
\def\tsc#1{\csdef{#1}{\textsc{\lowercase{#1}}\xspace}}
\tsc{WGM}
\tsc{QE}
\tsc{EP}
\tsc{PMS}
\tsc{BEC}
\tsc{DE}

\begin{document}
\let\WriteBookmarks\relax
\def\floatpagepagefraction{1}
\def\textpagefraction{.001}

\shorttitle{Similarity-Aware Multimodal Prompt Learning for Fake News Detection}

\shortauthors{Ye Jiang et~al.}

\title [mode = title]{Similarity-Aware Multimodal Prompt Learning for Fake News Detection}                      



%

\author[1]{Ye Jiang}[orcid=0000-0002-6683-0205]



\ead{ye.jiang@qust.edu.cn}


\credit{Conceptualization, methodology, writing - original draft}

\address[1]{School of Information Science and Technology, Qingdao University of Science and Technology, China}


\author[1]{Xiaomin Yu}[orcid=0000-0003-4846-3162]
\ead{yuxm02@gmail.com}
\credit{Software,  Writing - review \& editing}

\author[1]{Yimin Wang}[orcid=0000-0002-8835-3825]
\ead{yimin.wang@qust.edu.cn}
\cormark[1]
\cortext[cor1]{Corresponding author}
\credit{Project administration, writing – review \& editing}


\author[1]{Xiaoman Xu}[orcid=0000-0001-7448-4587]
\ead{xxm981215@163.com}
\credit{Data curation, writing – review \& editing}

\author[2]{Xingyi Song}[orcid=0000-0002-4188-6974]
\credit{Supervision, writing – review \& editing}
\ead{x.song@sheffield.ac.uk}
\address[2]{Department of Computer Science, University of Sheffield, United Kingdom}

\author[2]{Diana Maynard}[orcid=0000-0002-1773-7020]
\credit{Supervision, writing – review \& editing}
\ead{d.maynard@sheffield.ac.uk}
\begin{abstract}
Online fake news explosion has posed significant challenges to academics and industries by overloading fact-checkers and social media. The standard paradigm for fake news detection relies on utilizing text information to model news' truthfulness. However, the subtle nature of online fake news makes it challenging to use textual information alone to debunk it. Recent studies, focusing on multimodal fake news detection, have outperformed text-only methods. Deep learning approaches, primarily utilizing pre-trained models, to extract unimodal features, or fine-tuning the pre-trained model, has become a new paradigm for detecting fake news. Nevertheless, this paradigm may require a large number of training instances or updating the entire set of pre-trained model parameters, making it impractical for real-world fake news detection. In addition, traditional multimodal methods directly fuse the cross-modal features without considering that the uncorrelated semantic representation may introduce noise into the multimodal features. To address these issues, this paper proposed the \textbf{S}imilarity-\textbf{A}ware \textbf{M}ultimodal \textbf{P}rompt \textbf{Le}arning (SAMPLE) framework. Incorporating prompt learning into multimodal fake news detection, we used three prompt templates with a soft verbalizer to detect fake news. Additionally, we introduced the similarity-aware fusing method, which adaptively fuses the intensity of multimodal representation and mitigates noise injection via uncorrelated cross-modal features. Evaluation results show that SAMPLE outperformed previous works by achieving higher F1 and accuracy scores on two benchmark multimodal datasets, demonstrating its feasibility in real-world scenarios, regardless of data-rich or few-shot settings.

\end{abstract}



    
\maketitle

\section{Introduction}

\begin{figure}[h!]
	\centering
		\includegraphics[scale=.16]{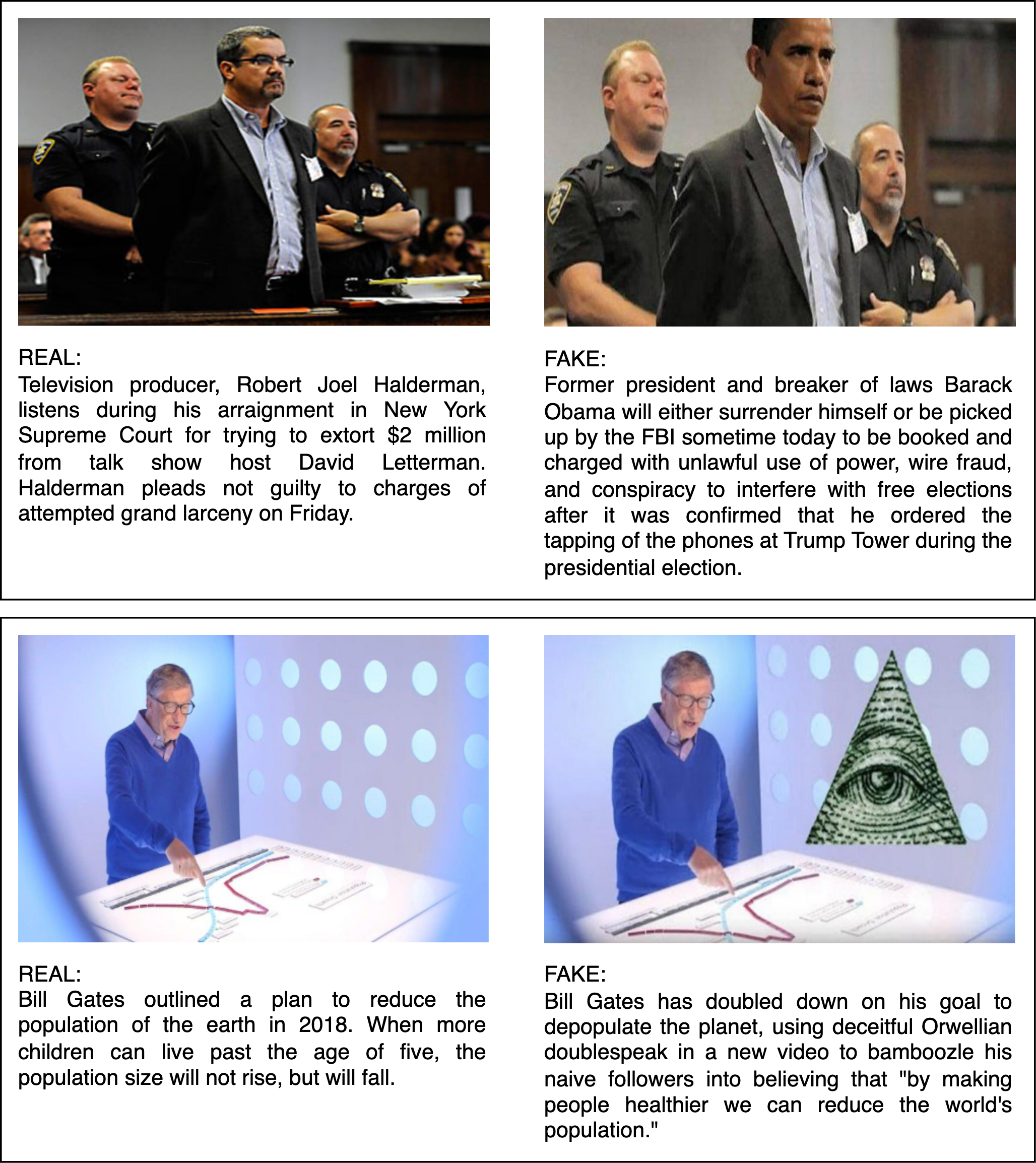}
	\caption{Two snippets of fake news and their original reports.}
	\label{FIG:fake}
\end{figure}

The increasing prevalence of social media has significantly impacted the way information is disseminated and consumed. While social media platforms provide an efficient way for people to seek and share information, the spread of fake news has caused substantial harm to the global community. In an effort to mitigate the impacts of online fake news, academia and industry have developed various techniques. For example, previous research \citep{ma2017detect, bahad2019fake, shu2019defend, dun2021kan} has mainly focused on the textual content of fake news. However, fake news can take various forms, and verifying its truthfulness by relying only on textual information requires expertise, which can be time-consuming. For example, Figure \ref{FIG:fake} shows two news snippets that are difficult to identify their truthfulness if only by looking at their text information. Therefore, multimodal Fake News Detection (FND) techniques \citep{wang2018eann, khattar2019mvae, chen2022cross} have been developed recently to combine both image and textual information, demonstrating promising performance as complementary benefits offered through cross-modality analysis. 

Multimodal FND aims to combine features from images and texts to automatically identify fake news posts. Traditional deep learning methods, such as Convolutional Neural Networks (CNNs), Recurrent Neural Networks (RNNs) and Transformers, have made significant advances in modelling both textual and image representations from fake news. However, these methods are limited by the fact that they often need a considerable amount of annotated data to achieve good performance. Recently, there has been an increasing interest in FND for using large pre-trained models. Much research \citep{singhal2019spotfake, singhal2020spotfake+, bhatt2022fake, zhou2020mathsf} has used pre-trained language models, such as BERT \citep{jacob2018} and RoBERTa \citep{liu2019roberta}, and pre-trained vision models, such as  ResNet \citep{he2016deep} and Vision-Transformer \citep{dosovitskiy2020image}, to encode the textual and image features of the news posts, respectively. However, pre-trained models are typically trained on a large, unrefined corpus that is not specific to any particular domain. Although pre-trained models can leverage external knowledge to identify fake posts, the effectiveness of an FND system is highly dependent on its focus domain \citep{jiang2022fake}.

Fine-tuning is a common technique for adapting pre-trained models to different downstream tasks. Recent studies have already fine-tuned variants of BERT, including the pre-trained BERT itself, for use in FND \citep{aggarwal2020classification, chen2021transformer, kumar2021fake}. However, fine-tuning requires a significant number of labeled instances to train additional classifiers, making it difficult to use in low-resource settings \citep{brown2020language}. Traditional pre-trained language models are trained with a cloze-style objective, which involves predicting masked words to learn their distributions, while fine-tuning aims to identify the target label directly. Consequently, pre-trained models require a significant amount of labeled data to be fine-tuned for specific tasks. Additionally, fine-tuning updates all model parameters for a single task, creating challenges for real-world FND due to the size of pre-trained models \citep{liu2022p}. Prompt learning is an approach that aims to better utilize pre-trained knowledge by adding additional information to the input and using a cloze-style task during the tuning process, resulting in more effective application of pre-training information \citep{schick2020exploiting}. Furthermore, prompt learning allows pre-trained models to achieve competitive performance in low-resource settings with limited labeled data \citep{brown2020language}, which is particularly important for real-world FND, where manually labeled fake news is scarce. However, current prompt-based FND approaches \citep{jiang2022fake} primarily consider textual information, and the analysis of cross-modality features in fake news posts is underdeveloped.

Compared to the fine-tuned model that outputs class distributions directly, prompt learning aligns with the language modeling objective, which generates specific answering words that are relevant to fake news detection by adding additional information before the original text inputs. For example, the news snippet on the left in Figure \ref{FIG:fake} can utilize a discrete prompt by adding a prompt before the original text (e.g., \textit{"This is a piece of <mask> news. Former president and breaker of laws, Barack Obama..."}), with the goal of recovering the masked token from the prompt text. However, the discrete prompt has limitations since the embedding of template words must be the embedding of natural language words, and the template can only be parameterized by the pre-trained LM parameters instead of the LM parameters that can be tuned via a specific task such as fake news detection \citep{lester2021power, liu2021gpt}. To address this issue, continuous prompting \citep{qin2021learning, zhong2021factual} eliminates the constraint of the discrete prompt by performing prompting directly in a continuous space of the pre-trained model, for example, \textit{"<soft><soft>...<soft><mask>. Former president and breaker of laws, Barack Obama..."}, where each <soft> can be associated with a randomly initialized trainable vector. Additionally, instead of utilizing a fully learnable prompt template, a mixed prompt \citep{gu2022ppt, jiang2022fake} incorporates trainable vectors into a discrete prompt template (e.g., \textit{"<soft> This is a piece of <mask> news <soft>. Former president and breaker of laws, Barack Obama..."}), and demonstrates superior performance to using each prompt type individually.

Previous multimodal FND methods \citep{wang2018eann, singhal2019spotfake} aimed to enhance performance by directly fusing multimodal representations. However, combining solely image and text features cannot guarantee reliable information, as the veracity of news articles is not completely associated with image-text correlation. In such cases, the correlation between text and image features is less, resulting in the multimodal representation being noisy. Therefore, it is crucial for multimodal FND models to grasp the semantic correlation between different modalities and adaptively combine multimodal features to conduct accurate classification.

This paper proposes a \textbf{S}imilarity-\textbf{A}ware \textbf{M}ultimodal \textbf{P}rompt \textbf{Le}arning (SAMPLE) framework for FND. Three popular prompt learning methods (discrete prompting (DP), continuous prompting (CP), and mixed prompting (MP)) are systematically integrated into a soft verbalizer in the task of FND. In addition, the pre-trained model Contrastive Language-Image Pre-training (CLIP) \citep{radford2021learning} is applied to extract the text and image features, which are utilized to generate the multimodal representation. Meanwhile, the semantic similarity between the text and image features is also calculated to address the issue of uncorrelated semantic representation between image and text. To adjust the intensity of the aggregated multimodal representation, the semantic similarity is normalized. To assess the performance of the proposed SAMPLE framework, two domain-specific publicly accessible datasets, PolitiFact and GossipCop \citep{shu2018fakenewsnet}), are utilized. We compare SAMPLE with existing FND methods, as well as the standard fine-tuning method, under both few-shot and data-rich scenarios to simulate real-world FND settings. The experimental results demonstrate that SAMPLE significantly outperforms traditional deep learning and fine-tuned approaches in both macro-f1 and accuracy metrics, regardless of data-rich or few-shot scenarios.

The contributions of this paper are:
\begin{itemize}
    \item We propose a framework called SAMPLE that adaptively fuses multimodal features generated by the CLIP model with textual representation from a pre-trained language model, to assist prompt learning for detecting fake news.

    \item The proposed framework mitigates the issue of uncorrelated cross-modal semantics by adjusting the intensity of fused multimodal features using standardized cosine similarity generated from the pre-trained CLIP model.

    \item SAMPLE are evaluated on two benchmark multimodal fake news detection datasets and demonstrates that it outperforms previous approaches in both low-resource and data-rich scenarios.

\end{itemize}

\section{Related work}

In previous research, FND has been extensively examined  \citep{zhang2020overview}. Specifically, fake news is described as false information that is circulated under the guise of being genuine news for political or financial gain via news outlets or the internet \citep{shu2017fake, meel2020fake}. In addition, many recent studies aim to differentiate false content from similar concepts, such as misinformation \citep{song2020classification, jiang2021categorising} and disinformation \citep{li2022classifying}. However, misinformation is false information that results from blunders or cognitive biases, whereas disinformation is intentionally fabricated, and in both cases, the formats are not limited to news outlets \citep{meel2020fake}. In this paper, we propose SAMPLE, which focuses primarily on FND but has the potential to extend to detecting misinformation and disinformation.

\subsection{Unimodal fake news detection}

Early research on unimodal FND often uses handcrafted features to identify anomalies in a post's text or image. Traditional methods of image manipulation detection \citep{chen2021image} can effectively detect tampering of news images. These methods learn image forensic, semantic, statistical, and contextual features from fake news. Fake news is frequently characterized by semantic inconsistencies that violate common sense \citep{li2021entity}, as well as poor image quality \citep{han2021fighting}. In text modality, textual features are commonly coupled with social features based on statistical information to detect fake news content in the news description \citep{castillo2011information, kwon2013prominent}. Previous research \citep{ajao2019sentiment, zhang2021mining} has investigated the connection between a publisher's emotions and social sentiment, which is closely tied to the accuracy of the news. In addition, logical soundness \citep{guo2018rumor}, grammatical errors \citep{potthast2017stylometric}, and rhetorical structure \citep{conroy2015automatic} are critical elements of FND. While unimodal FND is a robust baseline for detecting fake news, the correlation and consistency of the modalities in FND are not well established.

\subsection{Multimodal fake news detection}

Previous studies in multimodal fake news detection (FND) have typically focused on two approaches: designing complex networks or utilizing pre-trained models as feature extractors. The EANN \citep{wang2018eann} combines textual and visual features using a multi-task learning framework, which is designed to handle event classification and fake news detection simultaneously. One key aspect of this framework is that the event classification task removes post-specific information from fake news, while keeping invariant features that are useful for identifying rumors. \cite{zhou2020mathsf} proposed the SAFE model, which uses the Image2Sentence model \citep{vinyals2016show} to convert images to text captions, and extends the Text-CNN model \citep{kim2014convolutional} to extract textual features from news descriptions. To detect fake news, the model computes the relevance between the textual and visual information using a slightly modified cosine similarity measure, which is then fed into a classifier. 

More recently, many studies have opted to utilize pre-trained models to extract textual and visual features in FND. For example, the CAFE \citep{chen2022cross} employs BERT and ResNet-34 \citep{he2016deep} as pre-trained models for encoding textual and visual features, respectively. Similarly, \cite{tuan2021multimodal} utilized pre-trained BERT and VGG-19 models for extracting unimodal textual and visual features, respectively, and then applied a scaled dot-product attention mechanism to fuse the multimodal features. \cite{qi2021improving} proposed entity-enhanced multimodal fusion framework to capture entity inconsistency, mutual enhancement, and text complementation in the fake news. \cite{zhou2022multimodal} proposed the FND-CLIP model, which extracts feature representations from images and text using a ResNet-based encoder, a BERT-based encoder, and two pairwise CLIP encoders simultaneously.

Moreover, some studies have found that fine-tuning pre-trained models can also yield competitive performance, rather than just using them as feature extractors. As an example, \cite{yang2019xlnet} fine-tuned the pre-trained XLNet model for multi-class and binary class FND. The Ro-CT-BERT \citep{chen2021transformer} expands the vocabulary with professional phrases and adapts the heated-up softmax loss for adversarial training to improve the model’s robustness. Although traditional multimodal FND methods are known for accurately detecting fake news, they typically require a large amount of human-annotated data to train models effectively. Furthermore, while detecting fake news at an early stage can minimize its pernicious effects \citep{shu2021early}, FND methods are still limited by the availability of human-annotated data.

\subsection{Prompt learning for fake news detection}

In recent years, prompt learning has emerged as a new paradigm in Natural Language Processing (NLP) and has demonstrated comparable performance to standard fine-tuning in various NLP tasks. For example, \cite{zhu2022prompt} developed PLST, which combines both text inputs and external knowledge from open knowledge graphs in short text classification tasks. \cite{han2022ptr} proposed the PTR model, which is designed for many-class text classification, and constructs prompts using logic rules that contain multiple sub-prompts. Prompt-based models have also been used to aid fake news detection (FND). For example, \cite{el2021detecting} utilized the prompt-based model from DistilGPT-2 in conjunction with multitask learning to detect coronavirus-related fake news in MediaEval-2021. Moreover, \cite{jiang2022fake} proposed KPL, which detects fake news by integrating external knowledge. Nevertheless, KPL relies on human-designed prompts and verbalizers, which can be time-consuming and potentially unreliable. Additionally, it is not yet well understood how the fusion of multimodal representations of news posts can enhance fake news detection.

\section{Methodology}

The proposed approach aims to identify the authenticity of news articles by utilizing both text and image. The main objective of multimodal FND is to assign a standard binary classification label of $y\in\{0, 1\}$, where 0 represents real news and 1 represents fake news, to a given news article that includes both text input $x = [w_1, w_2, ..., w_n]$ with $n$ words and image input $i=[i_1, i_2, ..., i_m]$ with $m$ images. To determine the image that is most relevant to a given news article's text, the pre-trained CLIP model is utilized to encode both the text representation and image representation separately. To achieve this, only the image with the highest cosine similarity to the text is kept while the rest are discarded.

\begin{figure}[h!]
	\centering
		\includegraphics[scale=.06]{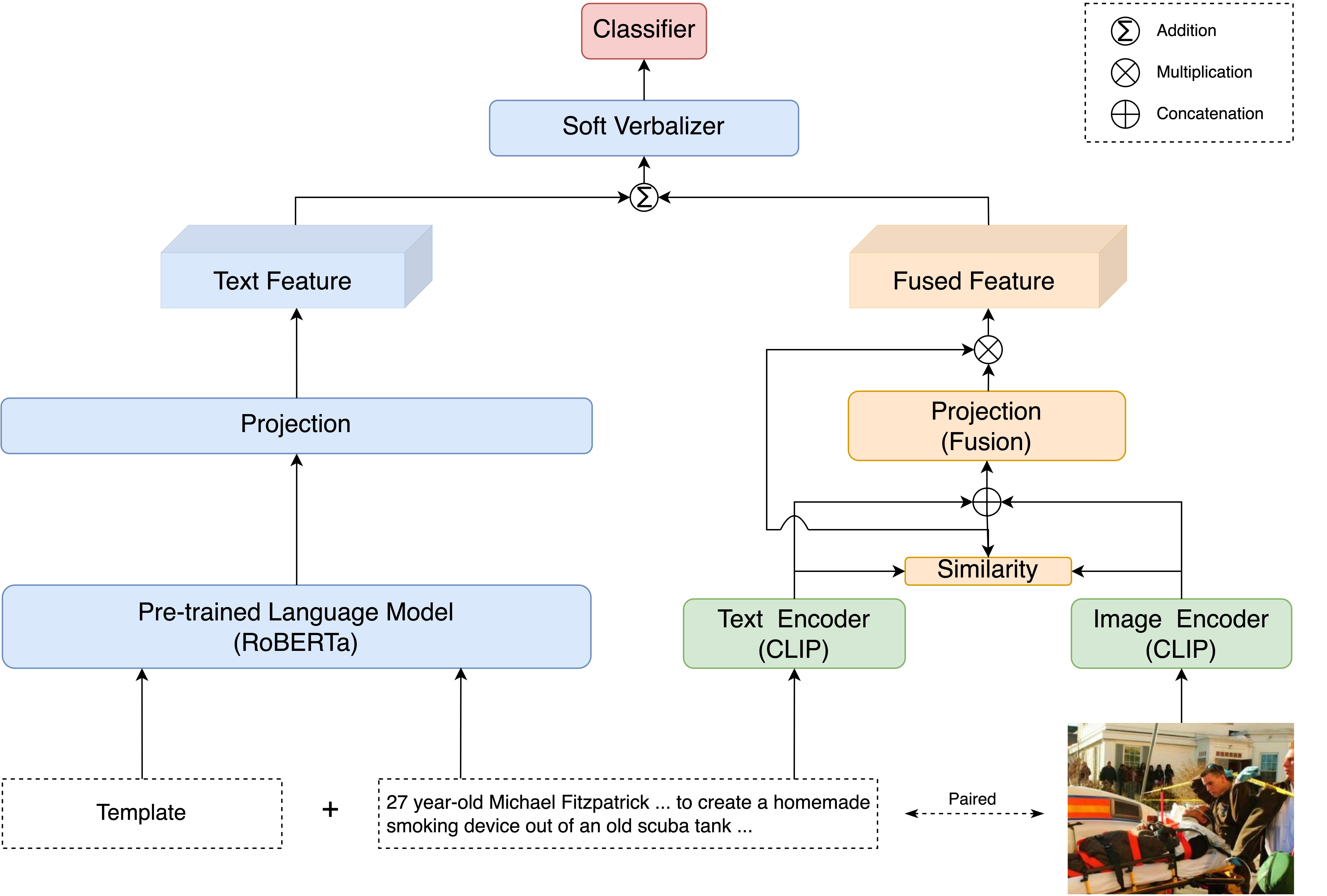}
	\caption{The overall structure of SAMPLE for fake news detection.}
	\label{FIG:overall}
\end{figure}

In this section, we utilize discrete prompting \citep{schick2021s, tam2021improving}, which primarily corresponds to natural language phrases and automatically searches for templates described in a discrete space. Additionally, we discuss the extended version called continuous prompting \citep{qin2021learning, zhong2021factual}, which employs prompts containing pseudo-tokens not present in the pre-trained language model vocabulary. We further present a mixed prompt that combines both discrete and continuous prompt for FND. Finally, we standardize the semantic similarity between the text and image to adjust the fused multimodal representation. Figure \ref{FIG:overall} illustrates the overall structure of the proposed SAMPLE.

\subsection{Discrete prompting}

We utilize a manually constructed discrete template as the prompting mechanism. To allow the model to retrieve the masked words, text inputs are initially masked during the prompt learning phase. The discrete prompting involves the intentional distortion of text input by means of a limited, human-designed template, with a singular keyword replaced with a mask. We investigate five discrete templates since the different templates might have a great impact on the performance of the language model as shown in Appendix \ref{app:1}. The discrete template $dt$ = ``This is a news piece with $<mask>$ information'', is the sum of the human-designed templates. After which, we can calculate the representation of the masked word, related to the FND task target, by the pre-trained language model. To accomplish this, we concatenate the discretionary template $dt$ with the initial input $x$ to generate the prompt, $x_d = [dt; x]$. Subsequently, we calculate the hidden states of prompt $x_d$:

\begin{equation}
    h_1^{dt}, ... , h_{mask}^{dt}, ... h_{m}^{dt} | h_1^{x}, ... , h_n^{x} = PLM(x_d)
\end{equation}
where $h^{dt}$ and $h_{mask}^{dt}$ are the hidden vectors in the $m$ length and the $<mask>$ token of discrete template respectively. $h^{x}$ are the hidden vectors of the $n$ length of the input text, and $PLM()$ is the masked language model output.

\subsection{Continuous prompting}

Although discrete prompting naturally inherits interpretability from the task description, the discrete prompt is limited because the embedding of template words is required to be natural language words, and the template is parameterized by the pre-trained language model's parameters \citep{liu2021pre}. In addition, discrete prompts may be suboptimal because the pre-trained language model may have learned the target knowledge from substantially different contexts. Such manually designed constraints can also be applied to the verbalizer because manual verbalizers usually determine predictions based on limited information. For example, the standard verbalizer maps fake $\longrightarrow$ \{counterfeit, sham,..., falsify\}, meaning that only predicting those related words for the token is considered correct during inference, regardless of the predictions for other relevant words like ``unreal'' or ``untrue'' that are also informative. Such a manually designed mapping limits the coverage of label words, resulting in insufficient information for prediction, and introducing bias into the verbalizer.

To address the above issues, the discrete template was reformatted by replacing trainable tokens with the continuous template $st =$ ``$<soft_1>,<soft_2>,...<soft_t>,<mask>$'' where each $<soft>$ is associated with a randomly initialized\footnote{We compares three initialization methods as shown in Appendix \ref{app:2}, and the experimental results suggest that the random initialization achieves comparable performance with a slightly faster convergence of validation loss than that of the others.} trainable vector. Then, the hidden states of the continuous prompt $x_s = [st; x]$ can be calculated similarly as:

\begin{equation}
    h_1^{st}, h_2^{st}, ... h_t^{st}, h_{mask}^{st} |  h_1^{x}, ... , h_n^{x} = PLM(x_s)
\end{equation}
where $h^{st}$ and $h_{mask}^{st}$ are the hidden vectors in the $t$ length and the $<mask>$ token of continuous template respectively.

\subsection{Mixed prompting}
Recent research has demonstrated that employing mixed prompting, which blends continuous and discrete templates, exhibits superior performance compared to using them independently \citep{liu2021gpt, han2022ptr}. Building on this, we have incorporated trainable tokens into the discrete prompt template. To be specific, we have inserted two trainable tokens, $h_{head}^{mt}$ and $h_{tail}^{mt}$, at the beginning and end of the mixed template $mt = $\textit{``$<h_{head}^{mt}>$ This is a piece of $<h_{mask}^{mt}>$ news. $<h_{tail}^{mt}>$ ''}.Like the discrete prompt, the new mixed prompt, $x_m$, can be expressed as $x_m = [mt; x]$. We then compute the hidden states as follows:

\begin{equation}
    h_{head}^{mt}, h_1^{mt}, ..., h_{mask}^{mt},...  , ... h_{m}^{mt}, h_{tail}^{mt} | h_1^{x}, ... , h_n^{x} = PLM(x_m)
\end{equation}
where $h^{mt}$ and $h_{mask}^{mt}$ are the hidden vectors in the $m$ length and $<mask>$ token of the mixed template respectively.


\subsection{Similarity-aware multimodal feature fusing}

According to a previous study \citep{zhou2022multimodal}, text and image features extracted from pre-trained models exhibit large semantic gaps. As a result, direct fusion of multimodal features fails to capture intrinsic semantic correlations. Unimodal pre-trained models, such as BERT and ViT-B-32, tend to focus on trivial clues, rather than on extracting semantically meaningful information. BERT can better learn emotional features from textual inputs, whereas ViT-B-32 can capture the noise patterns in images. Thus, direct fusion of unimodal features may inject noise into the multimodal representation, even if the text and the image are semantically correlated. Conversely, pre-trained CLIP models utilize a large dataset of image-text pairs to capture semantic correlations beyond emotional features or noise patterns.

To effectively integrate image and text features, we initially use the pre-trained CLIP model to extract these features separately. The CLIP model includes a text Transformer and the Vision Transformer (ViT-B-32) as the image encoder. To reduce the dimensionality of coarse features provided by the encoders and eliminate redundant information, we utilize individual projection heads $P_{txt}, P_{img}$ to process text and image features. Each projection head features two sets of fully-connected layers (FC), followed by Batch Normalization, a Rectified Linear Unit (ReLU) activation function, and a dropout layer. Next, we measure the cosine similarity $sim$ between $P_{txt}$ and $P_{img}$ to modify the intensity of the fused feature $f_{fused}$:

\begin{equation}
\begin{split}
        & f_{fused} = [P_{txt}; P_{img}]  \\
        & sim = \frac{P_{txt} (P_{img})^T}{||P_{txt}|| \; || P_{img}||} 
\end{split}
\label{eq:5}
\end{equation}

During the experiment, we observed that certain news posts lacked explicit cross-modal semantic relations, regardless of whether they were real or fake. As a result, concatenating the unimodal features to create the fused feature could add noise in instances where similarity was low. To remedy the issue, we apply standardization and a Sigmoid function to constrain the similarity value to the range of $[0-1]$. Standardization involves calculating the mean and standard deviation during training, subtracting the running mean from $sim$, and dividing the result by the running standard deviation. The standardized similarity can then be used to adjust the intensity of the final cross-modal representation, $m_{fused}$:

\begin{equation}
    m_{fused} = Sigmoid(Std(sim)) \, f_{fused}
\label{eq:6}
\end{equation}

\subsection{Soft verbalizer}

To recover masked words in the prompt template, the soft verbalizer is utilized to map labels to their corresponding words. We use WARP \citep{hambardzumyan2021warp} in this study to identify the optimal prompt for which a pre-trained language model predicts the masked token by searching for a prompt in the continuous embedding space. We use soft verbalization in the three template types above to compare prompt methods. To identify the optimal parameter $\theta = \{\theta^P, \theta^V\}$ for prompt and verbalizer embeddings, we first add the output vectors from the masked language model to the adjusted cross-modal representation using a residual connection. This combined output is then fed into an FC layer:

\begin{equation}
    x_{fused} = FC(PLM(x') + m_{fused})
\end{equation}
where $x'$ is the input sequence concatenate with one of the prompt template from the above, and then the classification probability $P(y|x')$ can be calculated as:

\begin{equation}
     P(y|x_{fused})  = \frac{\exp \theta_{y}^{V} x_{fused}}{  \sum_{i \in C} \exp \theta_{i}^{V} x_{fused}  }
\label{eq:3}
\end{equation}
where $C$ is the set of classes, $\theta_{y}^{V}$ is the embeddings of the true label and $\theta_{i}^{V}$ is the embeddings of the predicted label word. Finally, the cross-entropy loss can be minimized as:

\begin{equation}
    \theta^{*} = \arg \max_{\theta} (- \log P(y|x_{fused})) 
\end{equation}

\section{Experiment}
We evaluate our proposed approach on two benchmark FND datasets in low-resource and data-rich scenarios. The first part of this section presents an overview of the benchmark multimodal FND datasets, including their statistics. In the second part, we explain the implementation details for both the data-rich and few-shot settings. Finally, we provide a detailed discussion and analysis of our proposed method as well as the baseline models.

\subsection{Data}
We use two publicly accessible datasets for detecting fake information, namely PolitiFact and GossipCop, which consisted of political news and celebrity gossip, respectively, and are included in the FakeNewsNet project \citep{shu2018fakenewsnet}. Using the data crawling scripts provided, we retrieve 1,056 news items in PolitiFact and 22,140 news items in GossipCop. To reduce redundancy, we only preserve the relevant image, calculated by the pre-trained CLIP model based on the text and the images' cosine similarity, for news with multiple images. We also exclude news with no images or invalid image URLs. The resulting dataset statistics are presented in Table \ref{tab:1}.

\begin{table}[width=.7\linewidth,cols=4,pos=h]
\caption{The statistics of the pre-processed multimodal fake news datasets.}\label{tab:1}
\begin{tabular*}{\tblwidth}{@{} LLL @{} }
\toprule
Statistics  & PolitiFact & GossipCop \\
\midrule
Total news & 198 & 6,805 \\
Fake news & 96 & 1,877 \\
Real news & 102 & 4,928 \\
Words per news & 2,148 & 728 \\
\bottomrule
\end{tabular*}
\end{table}

\subsection{Implementation details}

We adopt the pre-trained RoBERTa from the HuggingFace library \citep{wolf2020transformers} as the main block for prompt learning. We extract text and image features using the text encoder and image encoder, respectively, from the pre-trained CLIP (ViT-B-32) model. The size of the hidden layer's projection layers is assigned as 768, and the dropout rate is 0.6. We use the AdamW optimizer to optimize the model parameters, and the learning rate $3\mathrm{e}{-5}$ and the decay parameter $1\mathrm{e}{-3}$ are empirically set. The model is trained in 20 epochs, and we choose the model checkpoints that obtain the best validation performance to test. We evaluate the method in both few-shot and data-rich settings.

In the few-shot setting, our model is trained with a small number of instances randomly sampled from the dataset, i.e., $k \in [2,4,8,16,100]$, and the rest of the instances are used for testing. Also, a validation set the same size as the training set is created for model selection. The PolitiFact dataset contains a limited number of news items, and to compensate for it, we adopt a special configuration called the PolitiFact 100-shot setting, where we use 100 instances for training and 50 for developing. As the quality of the training set and validation set has a significant impact on the model's performance, we repeat the above data sampling method five times with different random seeds and report the average score, excluding the highest and lowest ones, for the few-shot setting. For the few-shot setting, we report the average score of our model, computed by the mean of the scores without the maximum and the minimum ones. Additionally, we balance the number of labelled instances across the training and validation sets during the training phase.

In the data-rich setting, we split the two datasets into three parts, i.e., training set, validation set, and test set, with a split ratio of 8:1:1. In order to evaluate the stability of the proposed model, we repeat the above data sampling process five times with different random seeds. We report the average score, calculated as the mean of the scores after removing the highest and lowest ones from the five runs.

\subsection{Baseline models}

We compare the proposed SAMPLE to several models that have previously achieved state-of-the-art performances in FND dataset. Specifically, we compare unimodal approaches (1-2), multimodal approaches (3-6), and the standard fine-tuning approach (7). To initialize our word embeddings, we exploit the pre-trained 100-dim 6 billion Glove embeddings \citep{pennington2014glove}.

\begin{enumerate}
    \item[(1)] \textbf{LDA-HAN} \citep{jiang2020comparing}: This model incorporates Latent Dirichlet Allocation (LDA) \citep{blei2003latent} topic distributions into a hierarchical attention network.
    \item[(2)] \textbf{T-BERT} \citep{bhatt2022fake}: This feature-based method uses concatenated triple BERT models to predict fake news.
    \item[(3)] \textbf{SAFE} \citep{zhou2020mathsf}: This model converts images into their text descriptions and uses the relevance between textual and visual information to detect fake news.
    \item[(4)] \textbf{RIVF} \citep{tuan2021multimodal}: This model utilizes VGG and BERT models to encode image and text features. It applies the scaled dot-product attention mechanism on fused multimodal features to capture the relationship between text and images.
    \item[(5)] \textbf{SpotFake} \citep{singhal2019spotfake}: This model uses pre-trained image model VGG and BERT to extract respective image and text features, concatenating them to classify fake news.
    \item[(6)] \textbf{CAFE} \citep{chen2022cross}: This model uses an ambiguity-aware multimodal approach to adaptively aggregate unimodal features and correlations.
    \item[(7)] \textbf{FT-RoBERTa}: This is a standard, fine-tuned version of the pre-trained language model RoBERTa.

\end{enumerate}

\subsection{Results}

\begin{table}[width=\linewidth,cols=2,pos=t]
\caption{The overall macro-F1 and accuracy between baselines and the multimodal prompt learning framework. D-SAMPLE, C-SAMPLE and M-SAMPLE denote discrete prompting, continuous prompting and mixed prompting in the proposed SAMPLE framework respectively. }\label{tab:2}
\begin{tabular*}{\tblwidth}{@{} LLLLLLLLLLLLLL@{} }
\hline
Data &   Model & \multicolumn{10}{L}{Few shot (F1/Acc)} &  \multicolumn{2}{L}{Data rich(F1/Acc)} \\\cline{3-12}
     &       &  \multicolumn{2}{L}{2 } &  \multicolumn{2}{L}{4} &  \multicolumn{2}{L}{8} &  \multicolumn{2}{L}{16} & \multicolumn{2}{L}{100}  & & \\ 
\hline
\multirow{10}{4em}{PolitiFact} & LDA-HAN & 0.37 & 0.39 & 0.42 & 0.43 & 0.44 & 0.47 & 0.48 & 0.52 & 0.61 & 0.63  & 0.70 & 0.74 \\
 & T-BERT & 0.43 & 0.50 & 0.45 & 0.57 & 0.5 & 0.54 & 0.50 & 0.54 & 0.69 & 0.69 & 0.71& 0.75\\
 \cline{2-14}
  & SAFE & 0.19 & 0.19 & 0.21 & 0.21 & 0.29 & 0.27 & 0.33 & 0.49 & 0.46 & 0.56 & 0.64 & 0.65\\
 & RIVF & 0.35 & 0.48 & 0.43 & 0.51 & 0.42 & 0.48 & 0.40& 0.47 & 0.43 & 0.49 & 0.43 & 0.45\\
  & Spotfake & 0.37 & 0.49 & 0.46 & 0.52 & 0.47 & 0.54 & 0.56 & 0.59 & 0.73& 0.73 & 0.71 & 0.73\\
 & CAFE & 0.30 & 0.39 & 0.37 & 0.47 & 0.45 & 0.46 & 0.47 & 0.49 & 0.52 & 0.61 & 0.67 & 0.67 \\
  \cline{2-14}
   & FT-RoBERTa & 0.46 & 0.52 & 0.51 & \textbf{0.63} & 0.60 & 0.63 & \textbf{0.68} & \textbf{0.70} & 0.77 & 0.81 & 0.79 & \textbf{0.84}  \\
 \cline{2-14}
 & D-SAMPLE & 0.45 & 0.54 & 0.54 & 0.59 &0.61 & 0.64 & \textbf{0.68} & \textbf{0.70} & \textbf{0.81} & \textbf{0.82} & 0.79 & 0.81 \\
 & C-SAMPLE & \textbf{0.49} & 0.53 & 0.54 & 0.57 & 0.61 & 0.64 & 0.65 & 0.67 & 0.77 & 0.78 & \textbf{0.80} & 0.81\\
 & M-SAMPLE & 0.47 & \textbf{0.56} & \textbf{0.56} & \textbf{0.61} & \textbf{0.62} & \textbf{0.66} & 0.67 & \textbf{0.70} & \textbf{0.81} & \textbf{0.82} & \textbf{0.80} & 0.81\\
\hline
\hline
\multirow{10}{4em}{GossipCop} & LDA-HAN & 0.18 & 0.21 & 0.20 & 0.25 & 0.28 & 0.30 & 0.34 & 0.40 & 0.49 & 0.45 & 0.54 & 0.60\\
 & T-BERT & 0.38 & 0.48 & 0.38 & 0.57 & 0.45 & 0.66 & 0.45 & 0.71 & 0.52 & 0.61 & 0.61 & 0.74\\
 \cline{2-14}
  & SAFE & 0.26 & 0.31 & 0.33 & 0.41 & 0.40 & 0.45 & 0.41 & 0.45 & 0.44 & 0.51 & 0.55 & 0.64\\
 & RIVF & 0.24 & 0.29 & 0.24 & 0.29 & 0.24& 0.29 & 0.27 & 0.31 & 0.29 & 0.31 & 0.51 & 0.61 \\
  & Spotfake & 0.23 & 0.28 & 0.22 & 0.28 & 0.23 & 0.28 & 0.32 & 0.34 & 0.48 & 0.49 & 0.43 & 0.73\\
 & CAFE & 0.41 & 0.42 & 0.42 & 0.52 & 0.46 & 0.48 & 0.47 & 0.56 & 0.50 & 0.61 & 0.59 & 0.72 \\
  \cline{2-14}
  & FT-RoBERTa & 0.39 & 0.41 & 0.33 & 0.46 & 0.44 & \textbf{0.60} & 0.48 & \textbf{0.63} & 0.52 & \textbf{0.64} & 0.63 & 0.69  \\

  \cline{2-14}
  & D-SAMPLE & 0.42 & 0.47 & 0.44 & 0.50 & 0.50 & 0.58 & 0.51 & 0.59 & 0.57 & 0.62 & \textbf{0.64} & \textbf{0.76}  \\
 & C-SAMPLE & \textbf{0.47} & \textbf{0.54} & 0.46 & \textbf{0.56} & 0.45 & 0.52 & 0.46 & 0.53 & 0.52 & 0.58 & 0.63 & 0.75 \\
 & M-SAMPLE & 0.44 & 0.53 & \textbf{0.47} & \textbf{0.56} & \textbf{0.52} & 0.54 & \textbf{0.54} & 0.60 & \textbf{0.58} & 0.62 & \textbf{0.64} & 0.73\\
\bottomrule
\end{tabular*}
\end{table}
Table \ref{tab:2} shows the overall results that compare the proposed SAMPLE frameworks with fine-tuning approach, multimodal and unimodal FND methods.

\textbf{Comparing with fine-tuning.} First, we investigate the performances between the standard fine-tuned RoBERTa (FT-RoBERTa) and the proposed SAMPLE by evaluating the F1 score. We calculate the average improvements of M-SAMPLE (i.e., $\frac{(0.44-0.39)+...(0.58-0.52)}{5 \times 2} + \frac{(0.47-0.46)+...(0.81-0.77)}{5 \times 2}$), C-SAMPLE and D-SAMPLE to FT-RoBERTa, and all SAMPLE methods outperform FT-RoBERTa in 0.05, 0.024 and 0.035 respectively. This improvement is more significant with the decrease in the training samples, showing the superiority of prompt learning in low-resource scenarios.

However, the improvements become smaller in the data-rich setting, in which the average improvements of F1 are 0.005, 0.005 and 0.01 respectively, showing that the FT-RoBERTa is able to achieve comparable performance when the training data is sufficient. The comparison of accuracies between SAMPLE methods and FT-RoBETRa is similar to the above observation, showing the superiority of the proposed method in utilizing PLM information, especially when the training data is scarce, but the standard fine-tuning can still be a strong baseline in a data-rich setting. 

\textbf{Comparing with multimodal methods.}We evaluated the performance of SAMPLE in comparison with previous multimodal FND methods. Our results indicate that regardless of the multimodal and unimodal methods, both F1 and accuracy scores from SAMPLE outperformed previous methods in all settings. For example, with PolitiFact dataset, M-SAMPLE achieved a maximum 0.29 improvement in the 100-shot setting compared to CAFE. This observation is mainly attributed to the learning approach of the CLIP model that capitalizes on a large amount of image-text pairs to learn the extraction of multimodal semantics. In contrast, pre-trained models for unimodal tasks, such as BERT and ResNet-34 used by CAFE, might not be effective in capturing unimodal features with heterogeneous feature distributions.

The above characteristics are also applicable to SpotFake. SpotFake extracts text and image features using BERT and VGG19, respectively. However, our experiments demonstrate that SpotFake performs better on our smaller dataset, PolitiFact, compared to CAFE. This might be attributed to the fact that news topics in PolitiFact relate to politics, and hence, it is easier to fuse multimodal features by using pre-trained unimodality models without any ambiguous measurement. On the other hand, GossipCop presents a more complex semantic context as it consists of celebrity gossip stories. Therefore, CAFE's cross-modal ambiguity learning module performs better in handling the intricate cross-modal semantics of GossipCop.

\textbf{Comparing with unimodal methods.}In terms of unimodal methods, LDA-HAN exhibits a performance on par with multimodal methods when tested on Politifact, but not on GossipCop. This disparity could be due to the variance in context length between the two datasets, as revealed by the data in Table \ref{tab:1}. Specifically, PolitiFact contains a longer context length with 2,148 words per news than GossipCop, which contains only 728 words per news. Thus, the unimodal methods can extract more rich textual features from PolitiFact compared to GossipCop. Notably, although the unimodal T-BERT performs worse than the proposed SAMPLE, it demonstrates better performance than several multimodal methods in terms of F1 score and accuracy. We attribute this to the ensemble learning of T-BERT, which stacks three BERT models and shares the same weights during training. Despite the potentially higher computational cost associated with stacking multiple pre-trained language models, our findings also demonstrate the effectiveness of ensemble learning in FND.

\textbf{Analysis of different prompt templates.} The results indicate that mixed prompting (M-SAMPLE) outperforms C-SAMPLE and D-SAMPLE, with an average improvement of 0.04 and 0.02 in F1, respectively. This finding suggests that continuous prompting is inferior to the discrete and mixed prompting methods. Specifically, the use of the C-SAMPLE may not provide enough prior human knowledge to aid the verbalizer in capturing the label words from the continuous space.

Overall, the experimental results demonstrate that the proposed SAMPLE method exhibits superior performance in the task of multimodal FND, regardless of whether the few-shot or data-rich setting is employed.

\subsection{Analysis}
This subsection provides a thorough analysis of the proposed SAMPLE method in both few-shot and data-rich settings. First, we evaluate the significance of the image modality. Next, we present the standard deviations of the proposed model in various data settings. An ablation study further examines the key components of SAMPLE. Finally, we visualize and compare the embeddings from different baselines.

\subsubsection{Impact of image modality}
The integration of semantic similarity between image and text features into multimodal representation in SAMPLE enables automatic adjustment of relevance across multiple modalities. However, this method does not allow direct measurement of the effectiveness of the image modality.

\begin{figure}[h!]
     \centering
     \begin{subfigure}[b]{0.45\textwidth}
         \centering
         \includegraphics[width=\textwidth]{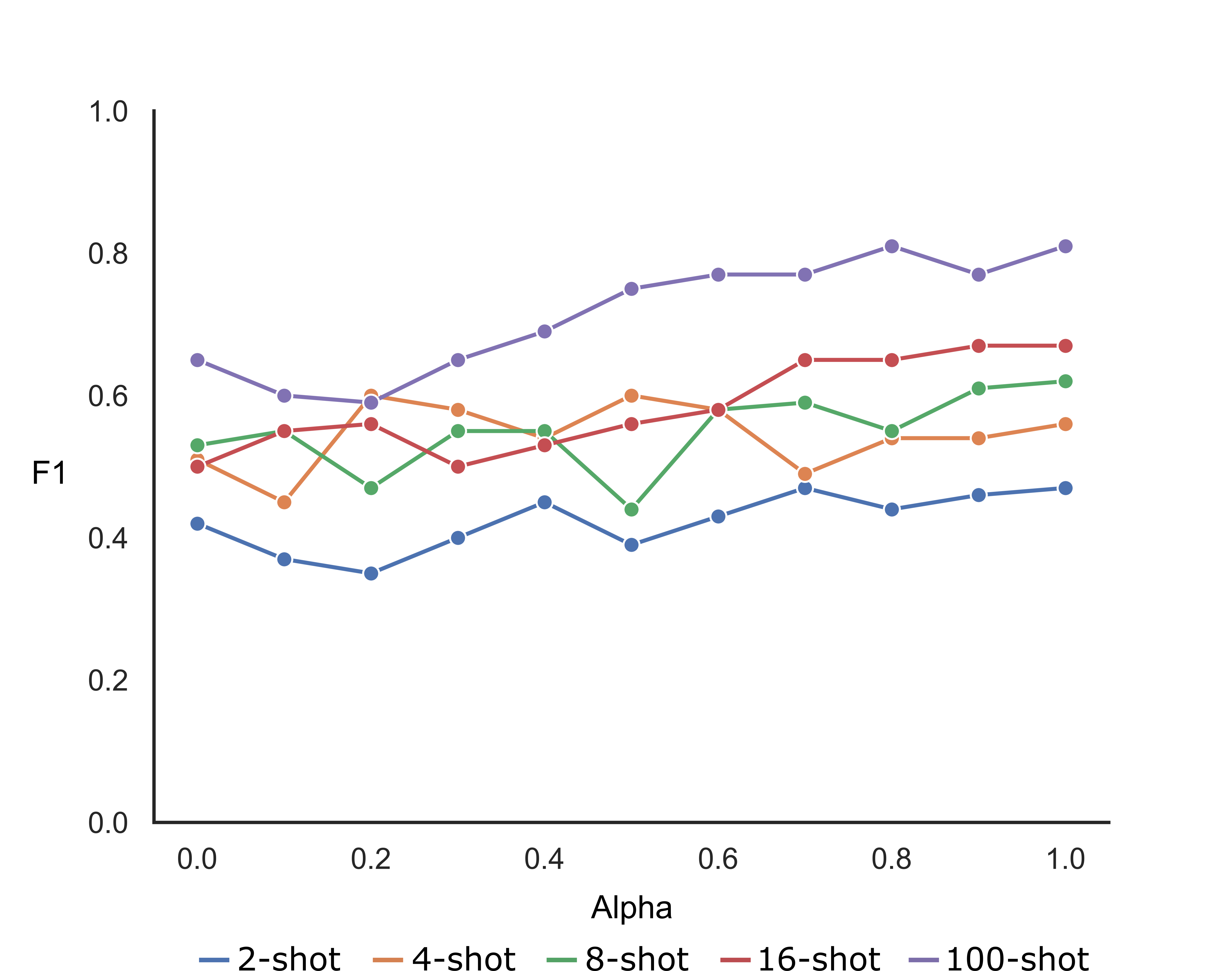}
         \caption{PolitiFact}
         \label{FIG:alpha}
     \end{subfigure}
     \begin{subfigure}[b]{0.45\textwidth}
         \centering
         \includegraphics[width=\textwidth]{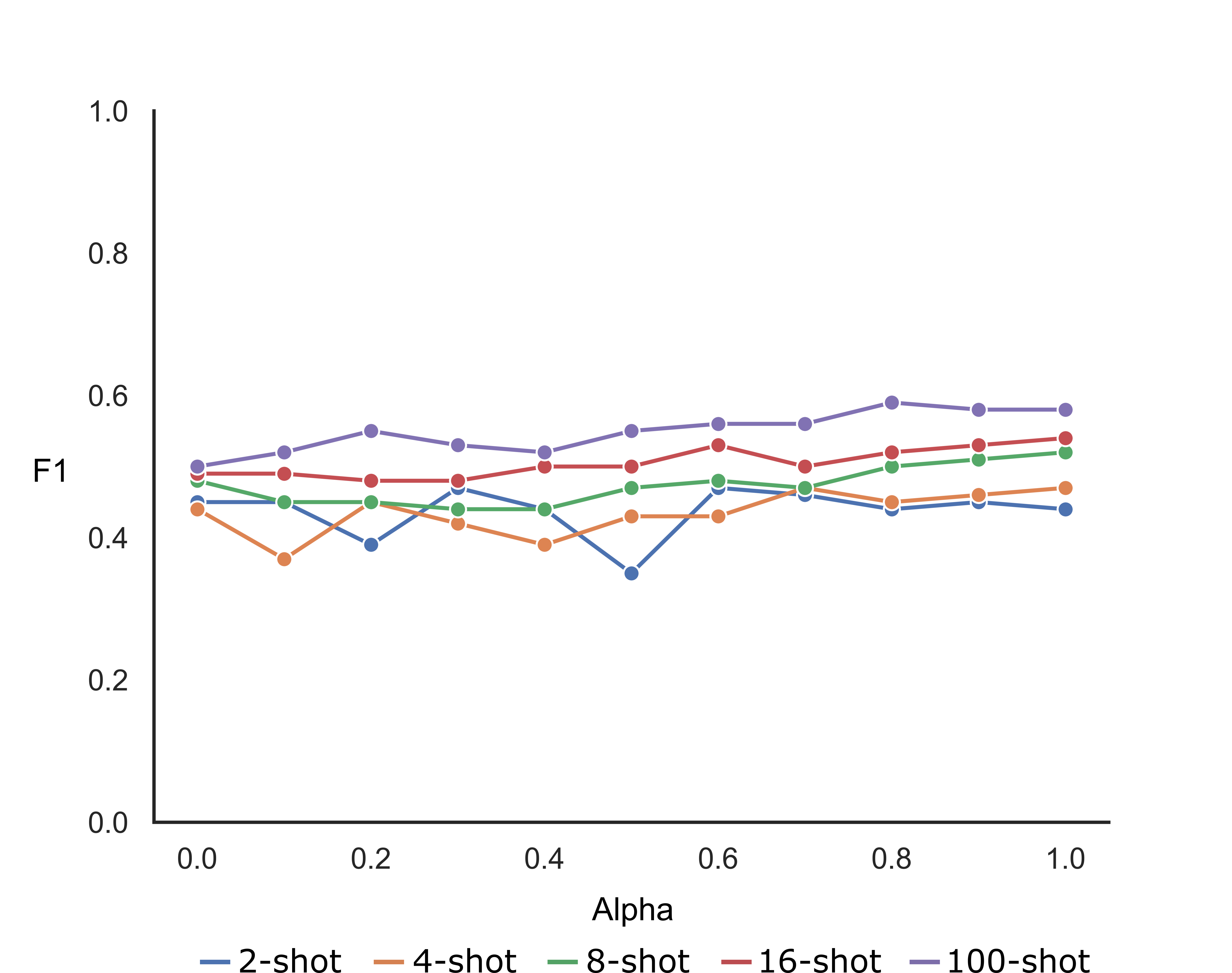}
         \caption{GossipCop}
         \label{FIG:alpha1}
     \end{subfigure}
        \caption{The importance of the image modality in the proposed framework.}
        \label{FIG:alpha_all}
\end{figure}

In order to comprehend the impact of the visual modality’s contribution to the model inference, we introduce an adjustable parameter, the parameter $\alpha$, to regulate the level of involvement of the visual modality in the few-shot training procedure. Precisely, the fused multimodal feature is multiplied by $\alpha \in [0,1]$. By setting $\alpha$ to 0, we eliminate the involvement of the visual modality. Conversely, if we set $\alpha$ to 1, both image and textual modalities are fully utilized. In this experiment, we apply M-SAMPLE, which achieves the highest F1. Based on the results depicted in Figure \ref{FIG:alpha_all}, M-SAMPLE attains a higher F1 as $\alpha$ increases, thereby indicating that the involvement of the visual modality can improve model performance. Nevertheless, we also observe that, in some instances, the inclusion of the visual modality leads to a decrease in the F1, especially when the number of training samples available is relatively small, such as in 2-shot, 4-shot, and 8-shot settings. This reflects that the features from the visual modality might harm the overall performance when there is less correlation with other modalities in the few-shot settings.

\subsubsection{Stability test}
\begin{figure}[h!]
     \centering
     \begin{subfigure}[b]{0.45\textwidth}
         \centering
         \includegraphics[width=\textwidth]{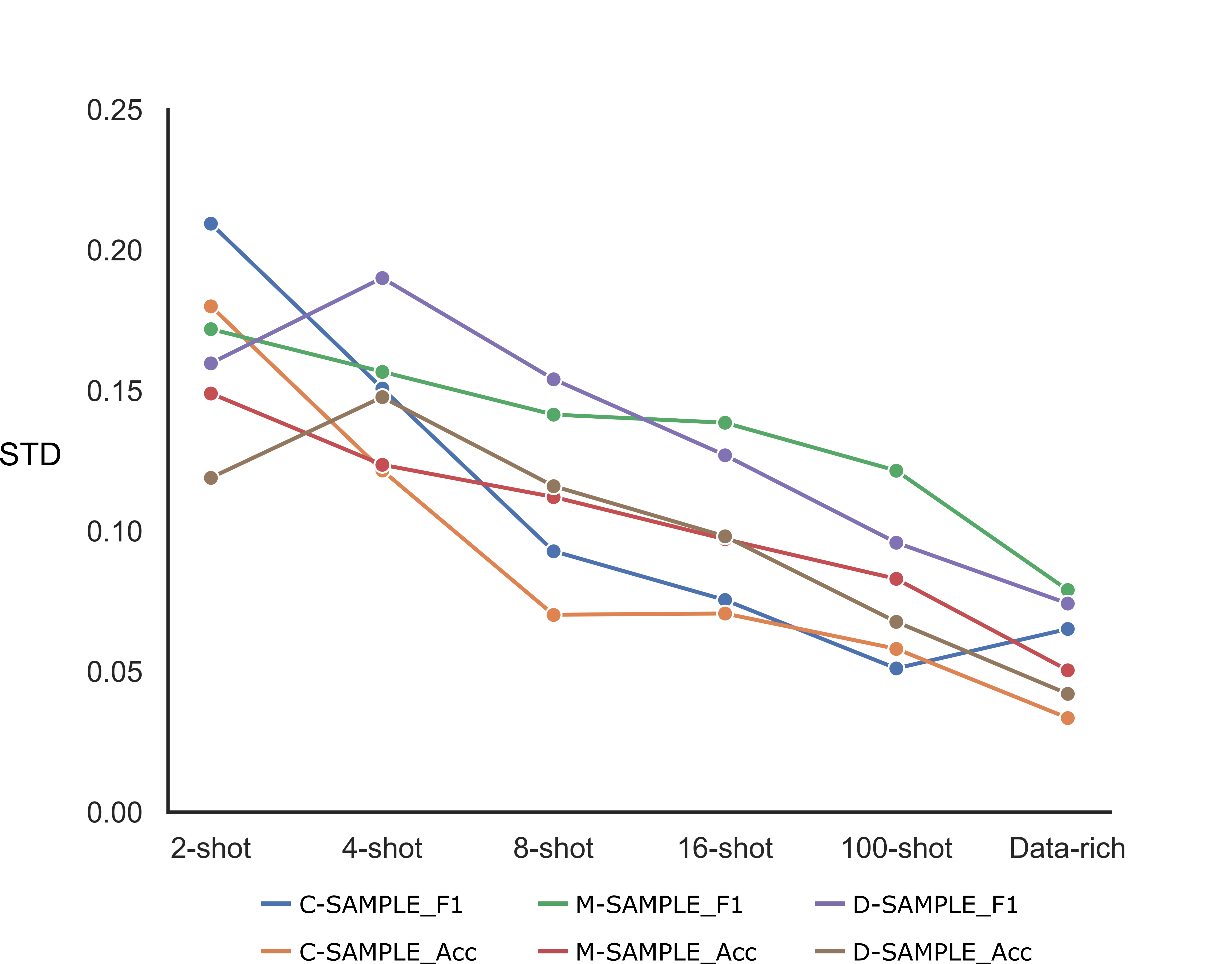}
         \caption{PolitiFact}
         \label{FIG:std1}
     \end{subfigure}
     \begin{subfigure}[b]{0.45\textwidth}
         \centering
         \includegraphics[width=\textwidth]{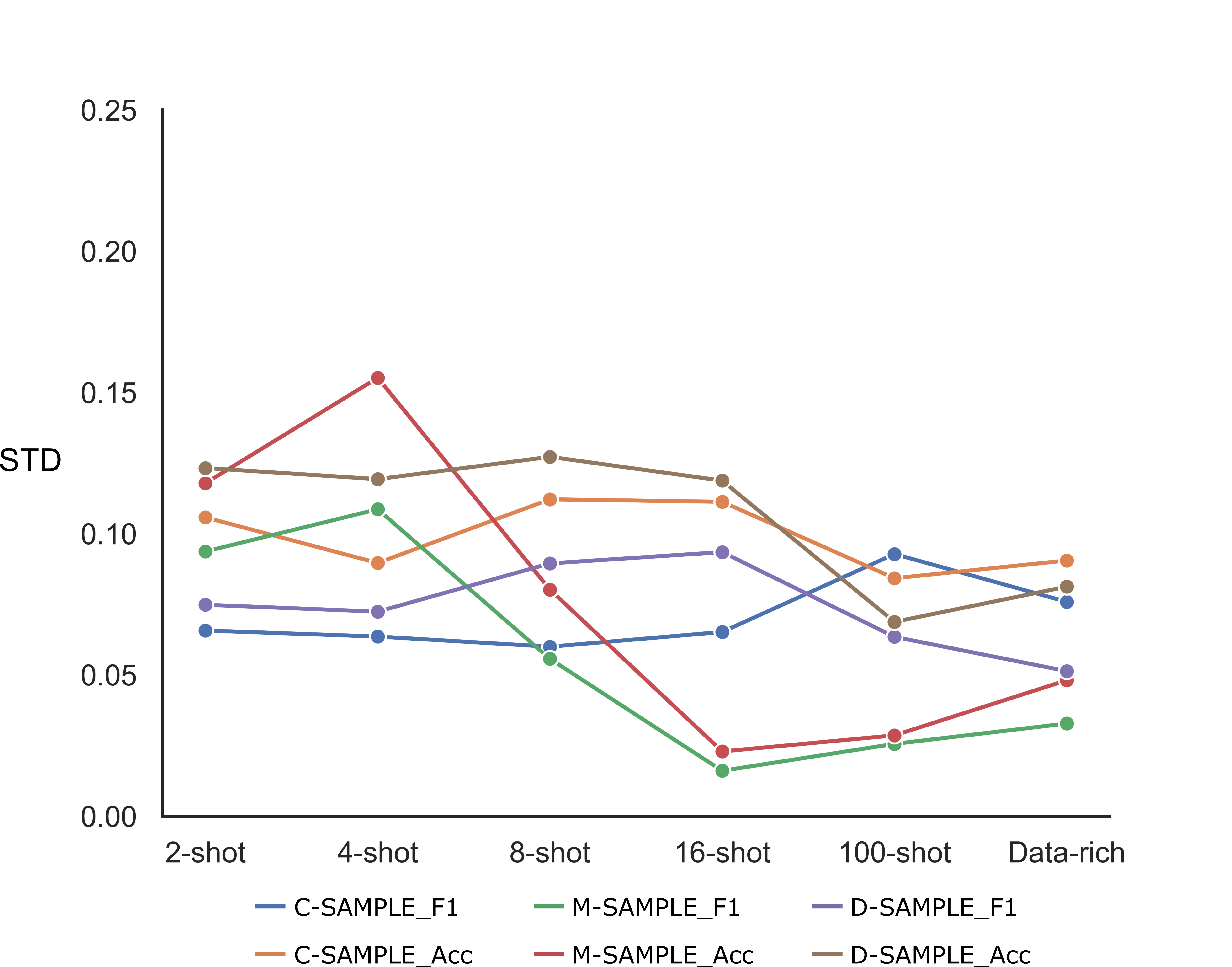}
         \caption{GossipCop}
         \label{FIG:std2}
     \end{subfigure}
        \caption{The standard deviation of the F1 and accuracy in the proposed framework.}
        \label{FIG:std_all}
\end{figure}

In this study, we evaluate the stability of the SAMPLE model by measuring the standard deviation of both F1 and accuracy in the few-shot and data-rich settings. As illustrated in Figure \ref{FIG:std_all}, we present the standard deviation of five experiments conducted for each SAMPLE model. We observe that the standard deviation decreases as the number of training samples increases, particularly in the PolitiFact dataset, as shown in Figure \ref{FIG:std1}. Moreover, the GossipCop dataset is relatively more unstable than The PolitiFact dataset, as shown in Figure \ref{FIG:std2}. This could be attributed to the complexity of semantics in GossipCop, which results in lower F1-score and accuracy for all models.

\subsubsection{Multimodal fusion strategies}
\begin{figure}[h!]
     \centering
     \begin{subfigure}[b]{0.45\textwidth}
         \centering
         \includegraphics[width=\textwidth]{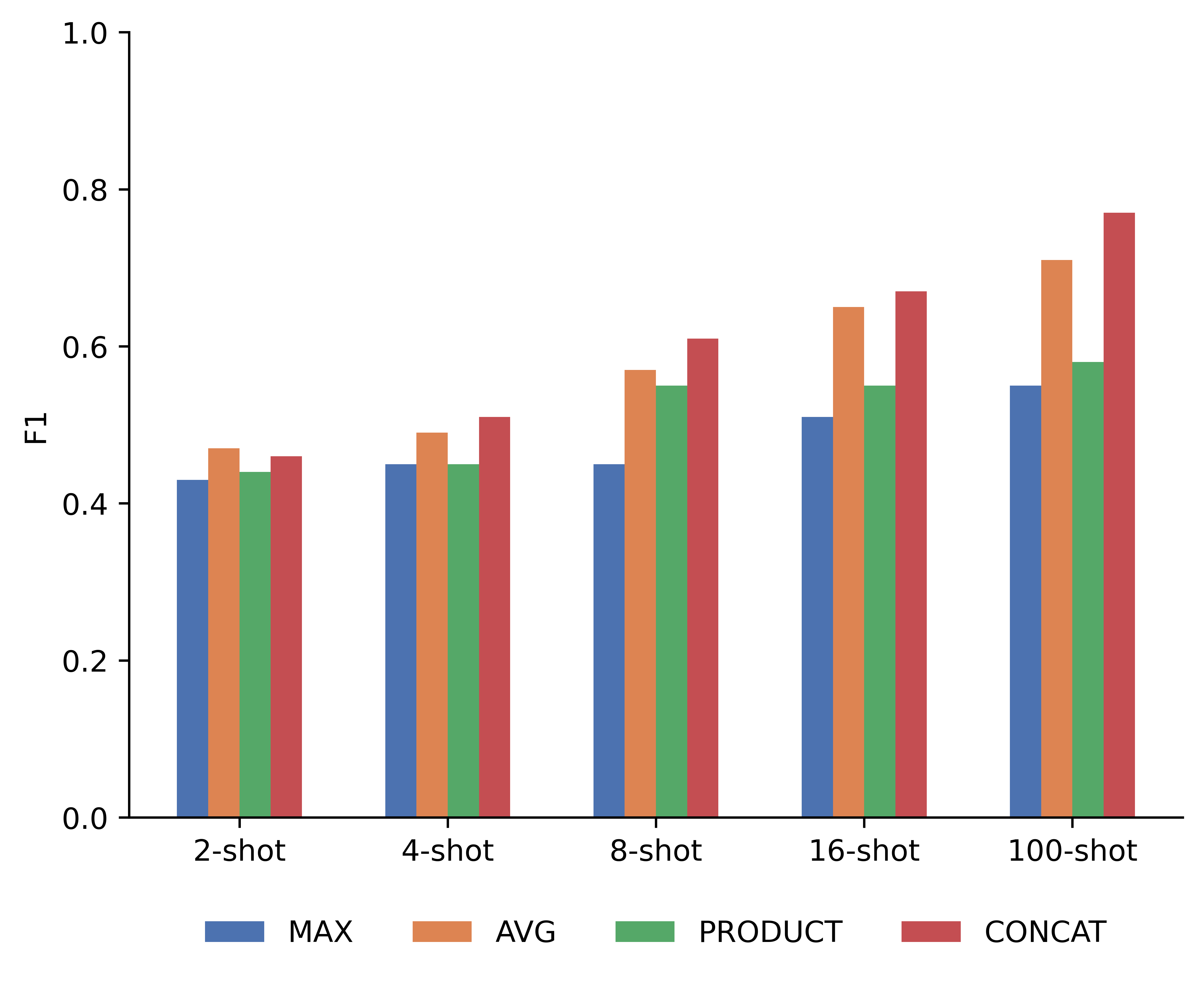}
         \caption{PolitiFact}
         \label{FIG:fusion1}
     \end{subfigure}
     \begin{subfigure}[b]{0.45\textwidth}
         \centering
         \includegraphics[width=\textwidth]{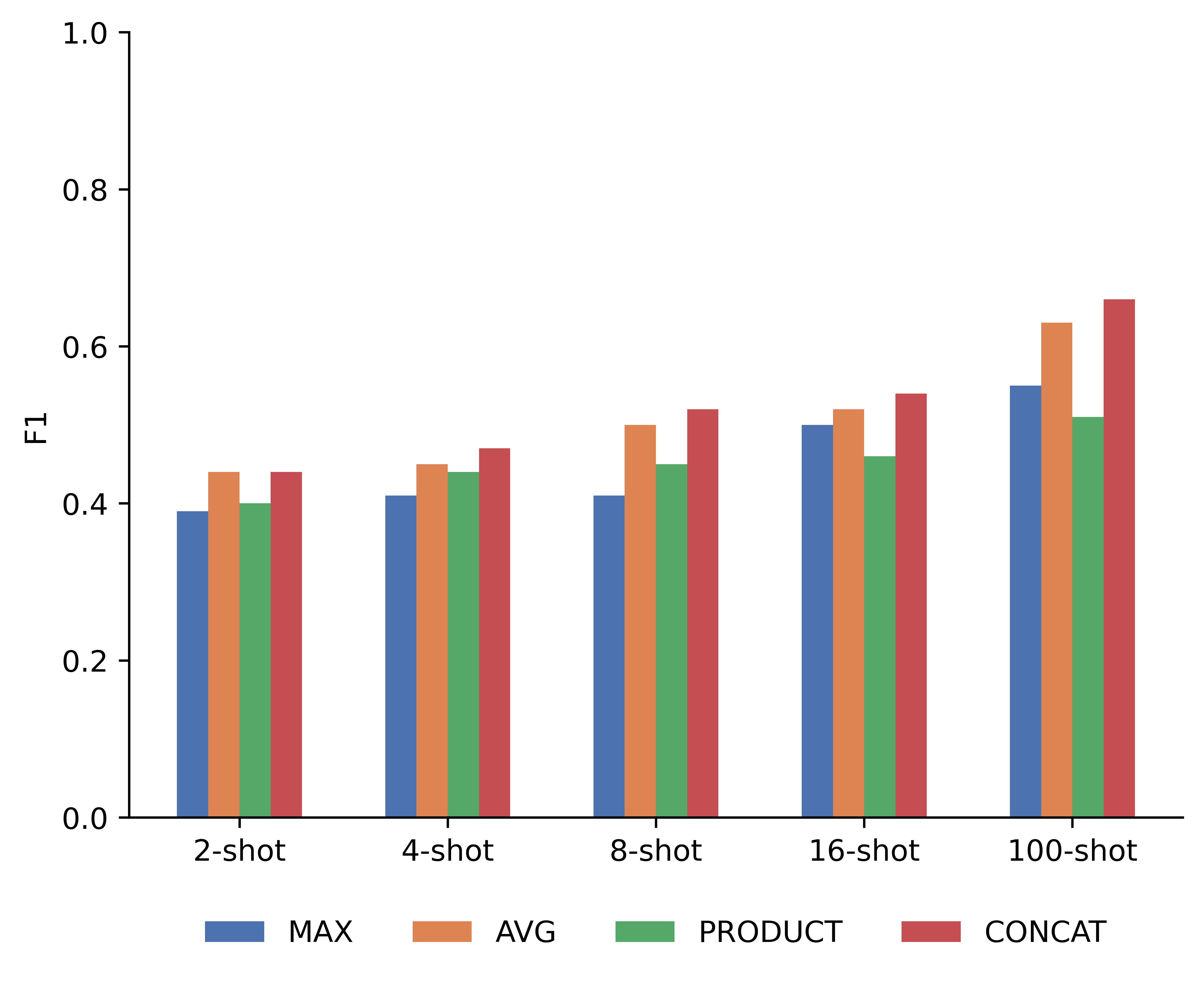}
         \caption{GossipCop}
         \label{FIG:fusion2}
     \end{subfigure}
        \caption{The comparison of different multimodal fusion strategies.}
        \label{FIG:fusion}
\end{figure}

The pre-trained clip model generates the unimodal features that normally come with better feature alignment than that generated from the unimodal pre-trained models \citep{zhou2022multimodal}. In this paper, we apply the M-SAMPLE and compare four rule-based fusion strategies\citep{atrey2010multimodal} to evaluate how the multimodal fusion would affect the performances in terms of F1 score, as shown in Figure \ref{FIG:fusion}. Specifically, ``MAX'' denotes the multimodal features are fused by a max-pooling layer, ``AVG'' denotes the multimodal features are fused by an average pooling layer, ``PRODUCT'' means multimodal features are made up of the output product of all unimodal features, ``CONCAT'' means the multimodal features are concatenated from the unimodal features. The results indicate that the concatenation of two unimodal features yields better f1 than other fusing strategies in the few-shot settings. 

\subsubsection{Trainable parameters comparisons}
We also compare the numbers of trainable parameters between baselines and SAMPLE as shown in Table \ref{tab:7}. The trainable parameters in SAMPLE frameworks are rather small and most of those come from the verbalizer and templates. In contrast, the FT-RoBERTa has the largest trainable parameter if fully fine-tuning the entire model. As a result, prompt learning is less computational cost than fine-tuning, and also can achieve comparable results compare with other deep learning methods.  

\begin{table}[width=.4\linewidth,cols=4,pos=h]
\caption{Trainable parameters between models. \#\_Para denotes trainable parameters in millions.}\label{tab:7}
\begin{tabular*}{\tblwidth}{@{} LL @{} }
\toprule
Model & \#\_Para   \\
\midrule
LDA-HAN & 0.17m   \\
T-BERT & 10m  \\
\hline
SAFE & 0.12m \\
RIVF & 8m \\
Spotfake &  13m \\
CAFE & 0.95m \\
\hline
FT-RoBERTa & 125m \\
\hline
D-SAMPLE & 0.64m \\
S-SAMPLE & 0.66m \\
M-SAMPLE & 0.64m \\
\bottomrule
\end{tabular*}
\end{table}

\subsubsection{Ablation study}

We investigate the influence of key components in SAMPLE by evaluating the framework's performance through various and partial setups. In each test, we employ M-SAMPLE, remove different components, and train the framework from scratch. The results are presented in Table \ref{tab:4} show that M-SAMPLE experiences a performance decay without each component in most setups, indicating the effectiveness of each key module in SAMPLE. Specifically, we observe a slight decrease in performance when removing the automatic similarity adjustment ``-SA''. This demonstrates that the standardizing of semantic similarity in fusing multimodal features helps reduce uncorrelated information in classifying fake news while mitigating the noise from different modalities' multimodal features.

\begin{table}[width=\linewidth,cols=2,pos=h]
\caption{The experimental results of ablation study based on M-SAMPLE. ``-SA'' denotes the automatic similarity adjustment is removed from M-SAMPLE, ``-IF'' means we remove the image feature from CLIP model, ``-TF'' means we remove the text feature from CLIP model, ``-MF'' means the multimodal feature from CLIP model is removed and only use the text feature from language model RoBERTa.}\label{tab:4}
\begin{tabular*}{\tblwidth}{@{} LLLLLLLLLLLLLL@{} }
\hline
Data &   Method & \multicolumn{10}{L}{Few shot (F1/Acc)} &  \multicolumn{2}{L}{Data rich(F1/Acc)} \\\cline{3-12}
     &       &  \multicolumn{2}{L}{2 } &  \multicolumn{2}{L}{4} &  \multicolumn{2}{L}{8} &  \multicolumn{2}{L}{16} & \multicolumn{2}{L}{100}  & & \\ 
\hline
\multirow{4}{4em}{PolitiFact}
                   & M-SAMPLE & 0.47 & 0.56 & 0.56 & 0.61 & 0.62 & 0.66 & 0.67 & 0.70 & 0.81 & 0.82 & 0.80 & 0.81\\
    & \multicolumn{1}{r}{-SA} & 0.44 & 0.55 & 0.55 & 0.57 & 0.58 & 0.65 & 0.65 & 0.65 & 0.75 & 0.79 & 0.76& 0.81\\
    & \multicolumn{1}{r}{-IF} & 0.43 & 0.53 & 0.51 & 0.57 & 0.55 & 0.63 & 0.60 & 0.61 & 0.73 & 0.77 & 0.75 & 0.78\\
    & \multicolumn{1}{r}{-TF} & 0.35 & 0.43 & 0.46 & 0.50 & 0.51 & 0.60 & 0.54 & 0.65 & 0.66 & 0.69 & 0.69 & 0.71\\
    & \multicolumn{1}{r}{-MF} & 0.32 & 0.48 & 0.43 & 0.51 & 0.46 & 0.56 & 0.55 & 0.57 & 0.63 & 0.69 & 0.65 & 0.65\\
\hline
\hline
\multirow{4}{4em}{GossipCop} 
                   & M-SAMPLE & 0.44 & 0.53 & 0.47 & 0.56 & 0.52 & 0.54 & 0.54 & 0.60 & 0.58 & 0.62 & 0.64 & 0.73\\
    & \multicolumn{1}{r}{-SA} & 0.43 & 0.50 & 0.45 & 0.57 & 0.50 & 0.51 & 0.50 & 0.59 & 0.55 & 0.69 & 0.59 & 0.75\\
    & \multicolumn{1}{r}{-IF} & 0.39 & 0.43 & 0.43 & 0.50 & 0.48 & 0.55 & 0.50 & 0.55 & 0.53 & 0.65 & 0.53 & 0.70\\
    & \multicolumn{1}{r}{-TF} & 0.37 & 0.49 & 0.38 & 0.49 & 0.41 & 0.50 & 0.44 & 0.50 & 0.49 & 0.56 & 0.51 & 0.65\\
    & \multicolumn{1}{r}{-MF} & 0.35 & 0.38 & 0.39 & 0.45 & 0.40 & 0.47 & 0.41 & 0.49 & 0.45 & 0.55 & 0.49 & 0.55\\
\bottomrule
\end{tabular*}
\end{table}

Furthermore, removing the text feature from CLIP ``-TF'' resulted in worse F1 and accuracy compare to the framework that remove the image feature from CLIP-``-IF'' in general. Our findings indicate that, although the image modality provides valuable information for FND (as shown in Figure \ref{FIG:alpha_all}), text features are still critical in prompt learning. This is mainly due to prompt learning's training objective which is to recover the masked token from the templates, primarily aligning text features extracted from pre-trained models. Additionally, as two text features are extracted from different pre-trained models (RoBERTa and CLIP, respectively), they give the classifier more expressive textual information. However, image features are mainly utilized to mitigate noise between different modalities.

We also removed fused multimodal features ``-MF'' generated from the CLIP model. In this partial setup, the proposed framework is the vanilla version of the prompt learning approach that leverages the pre-trained language model to predict FND directly. We discovered that vanilla prompt learning can still yield better performance than unimodal methods that utilize textual features only, highlighting the superiority of prompt learning in FND.

\begin{figure}[h!]
     \centering
     \begin{subfigure}[b]{0.24\textwidth}
         \centering
         \includegraphics[width=\textwidth]{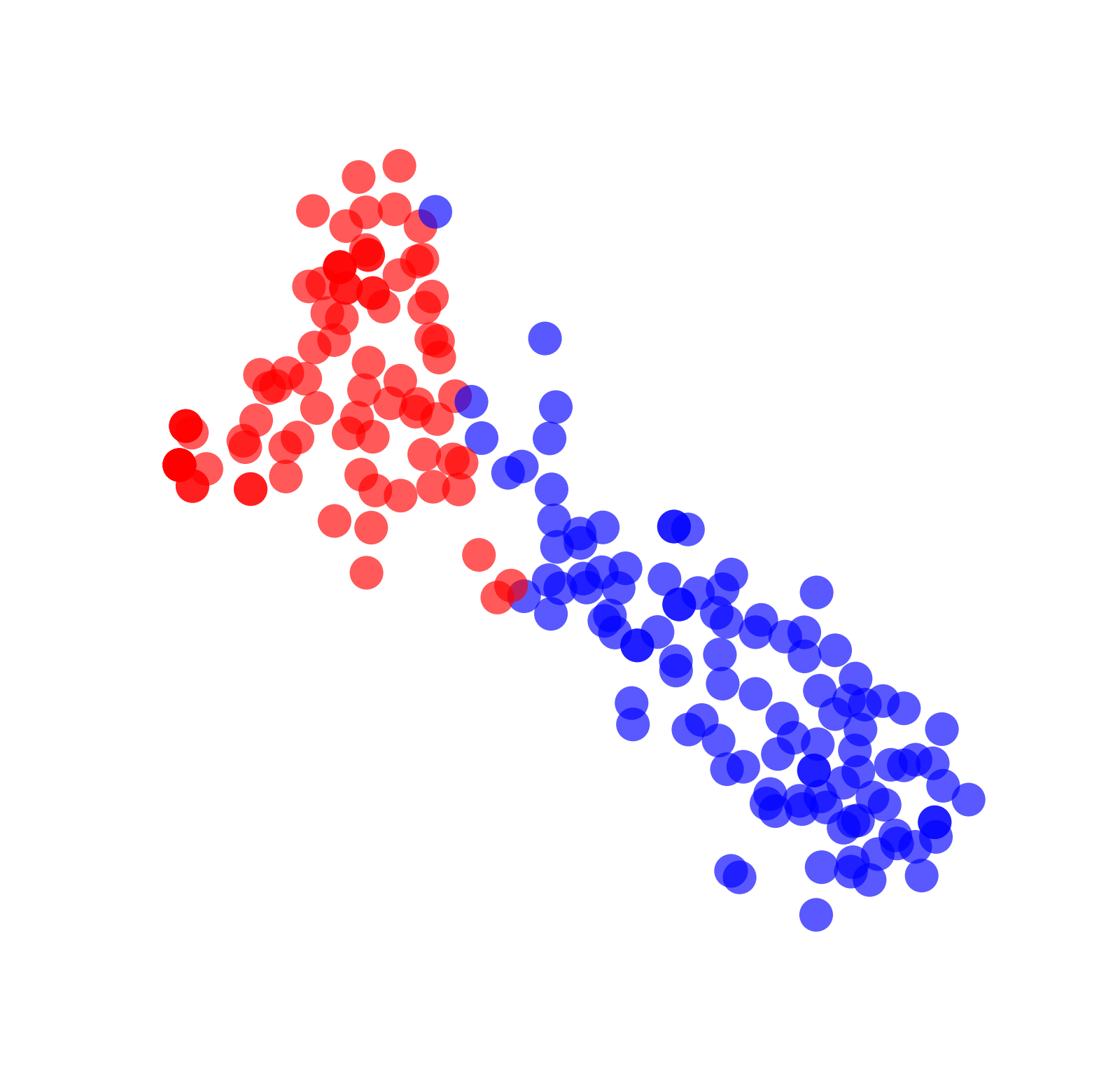}
         \caption{M-SAMPLE}
         \label{FIG:m-cloud}
     \end{subfigure}
     \begin{subfigure}[b]{0.24\textwidth}
         \centering
         \includegraphics[width=\textwidth]{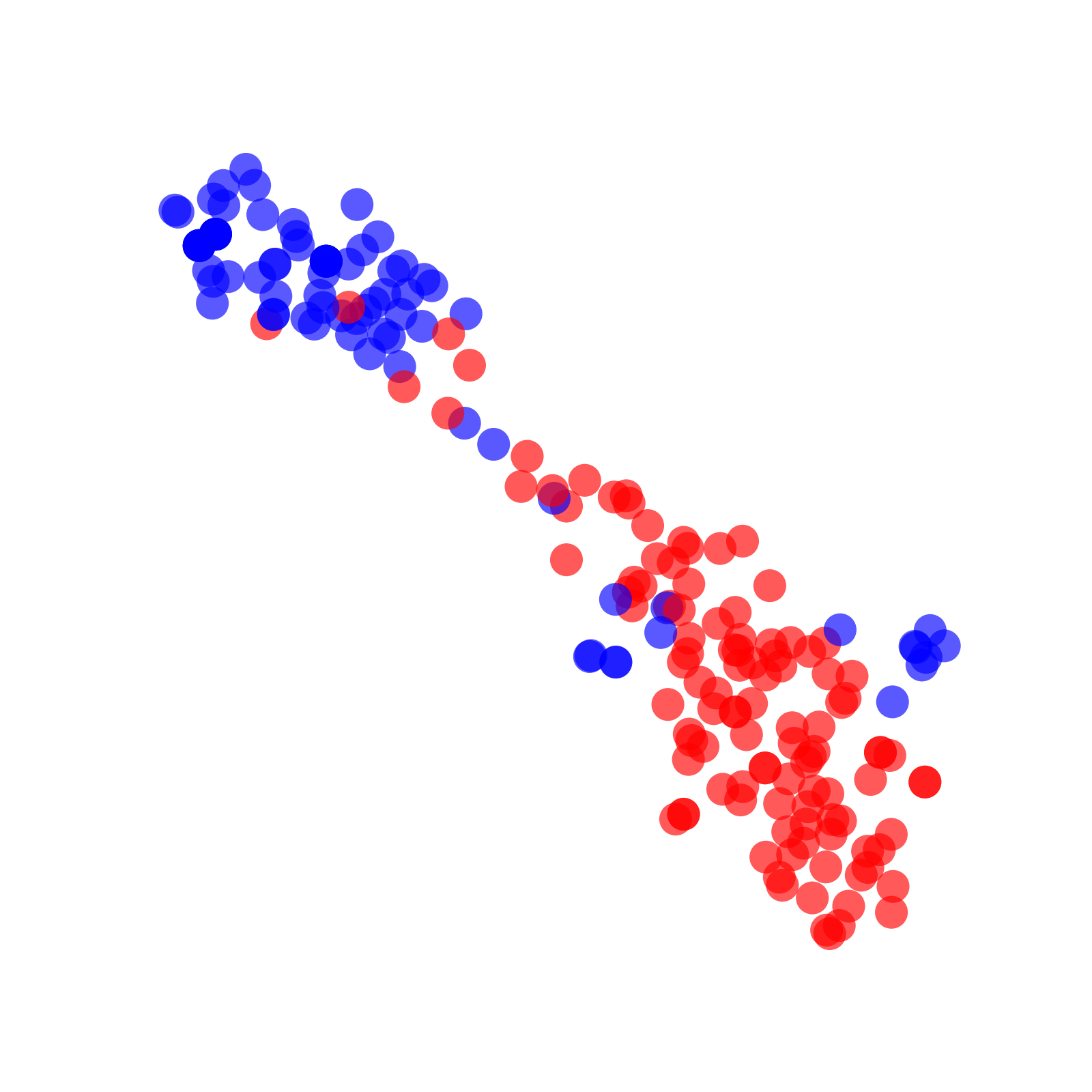}
         \caption{D-SAMPLE}
         \label{FIG:d-cloud}
     \end{subfigure}
     \begin{subfigure}[b]{0.24\textwidth}
         \centering
         \includegraphics[width=\textwidth]{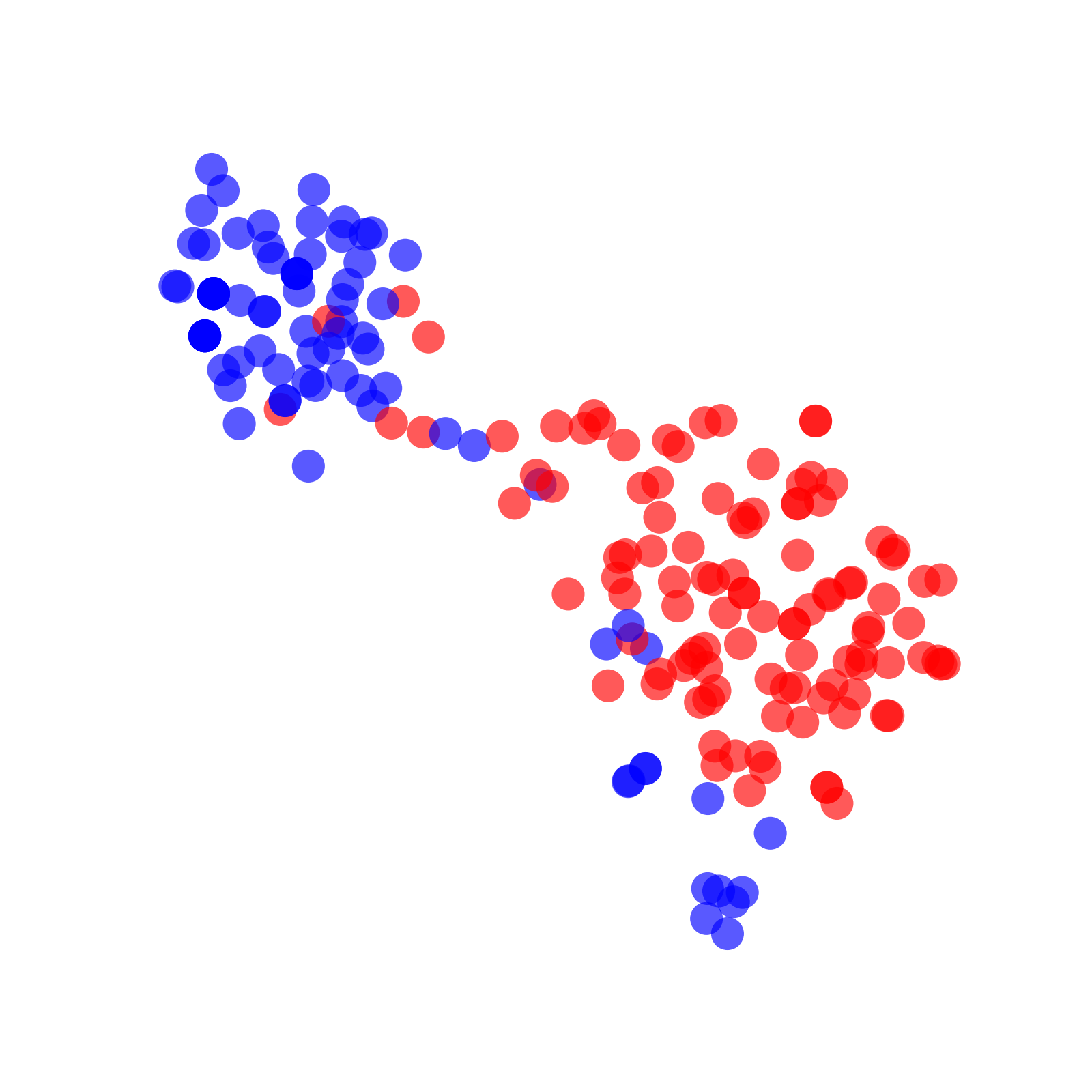}
         \caption{C-SAMPLE}
         \label{FIG:c-cloud}
     \end{subfigure}
     \begin{subfigure}[b]{0.24\textwidth}
         \centering
         \includegraphics[width=\textwidth]{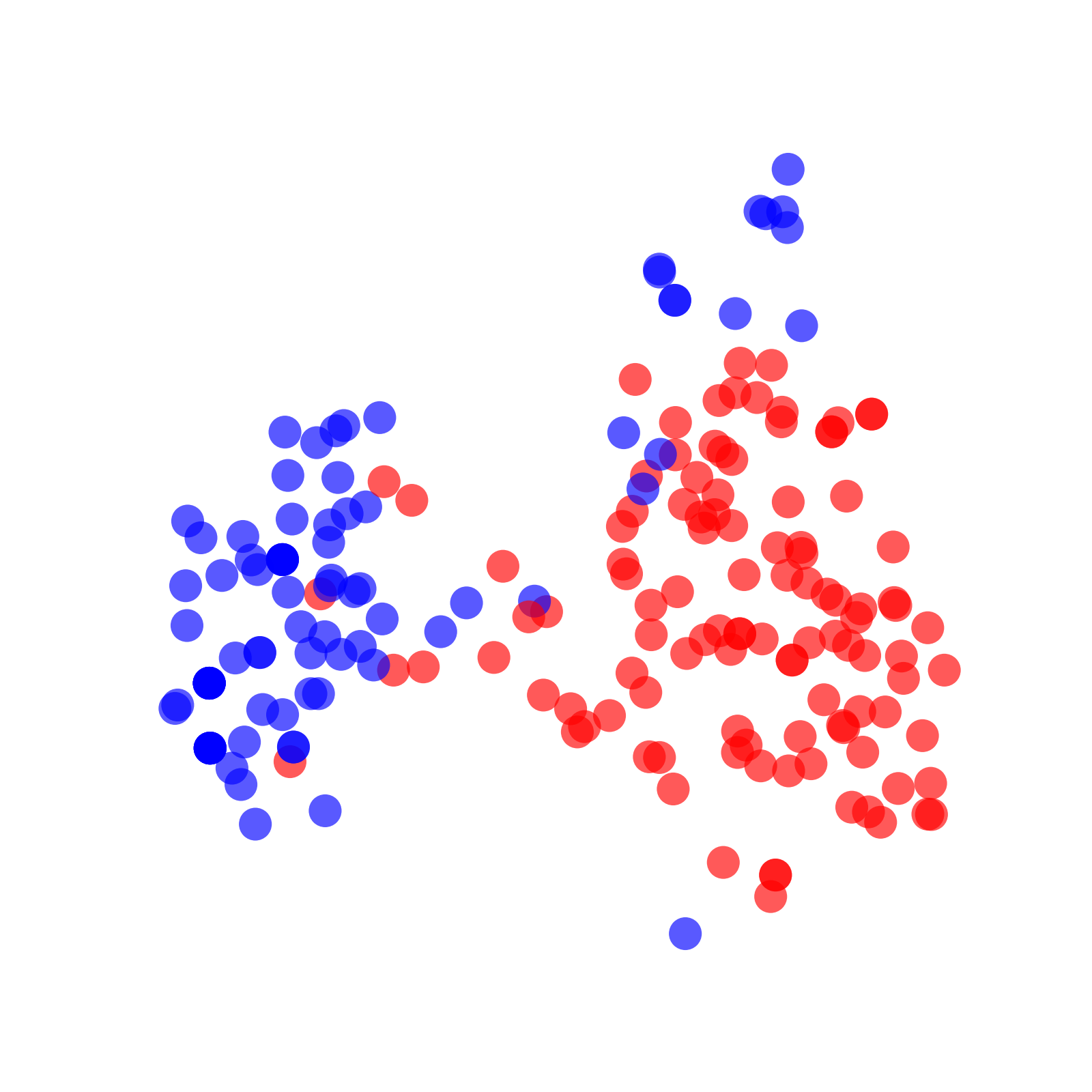}
         \caption{RoBERTa}
         \label{FIG:roberta-cloud}
     \end{subfigure}
     \begin{subfigure}[b]{0.24\textwidth}
         \centering
         \includegraphics[width=\textwidth]{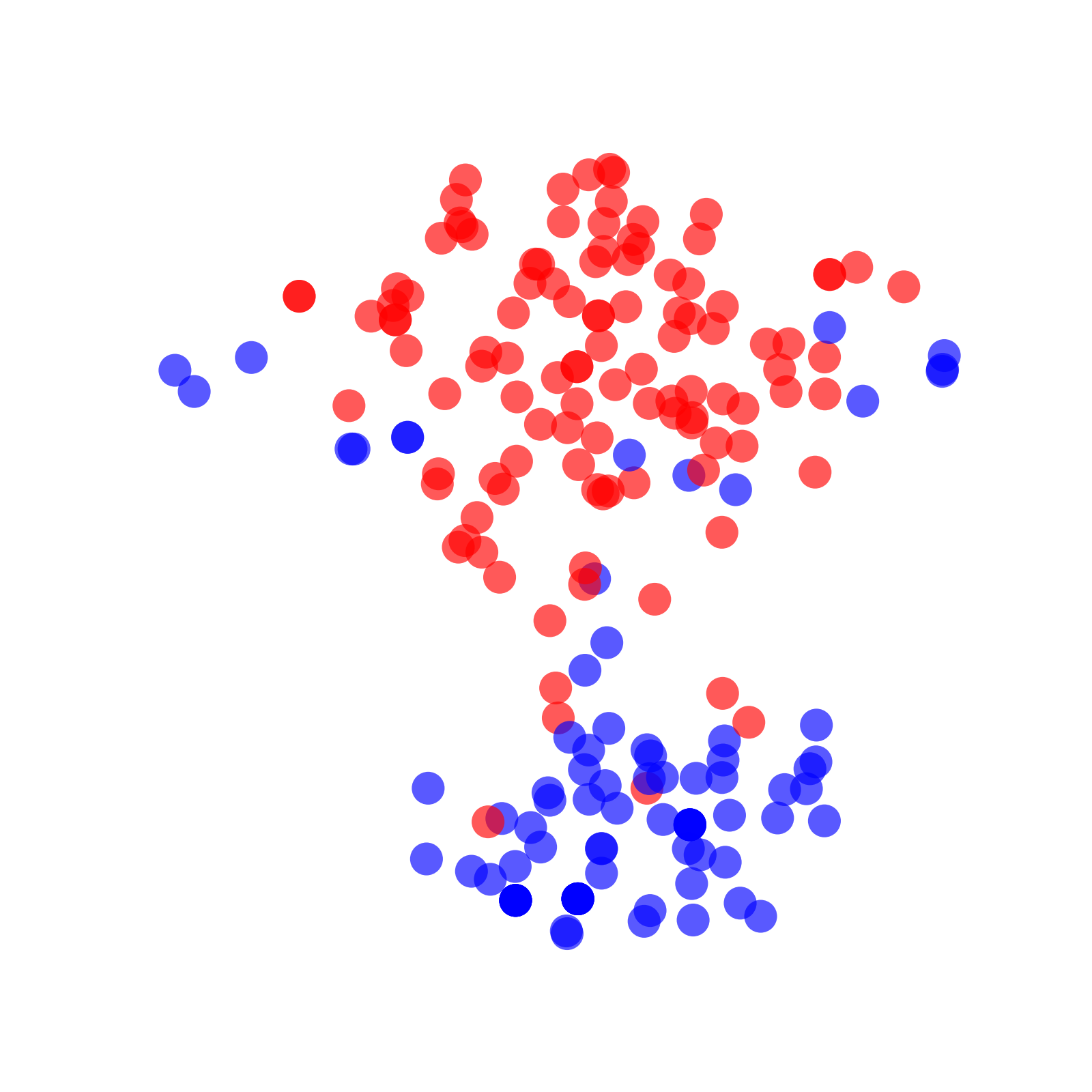}
         \caption{CAFE}
         \label{FIG:CAFE-cloud}
     \end{subfigure}
     \begin{subfigure}[b]{0.24\textwidth}
         \centering
         \includegraphics[width=\textwidth]{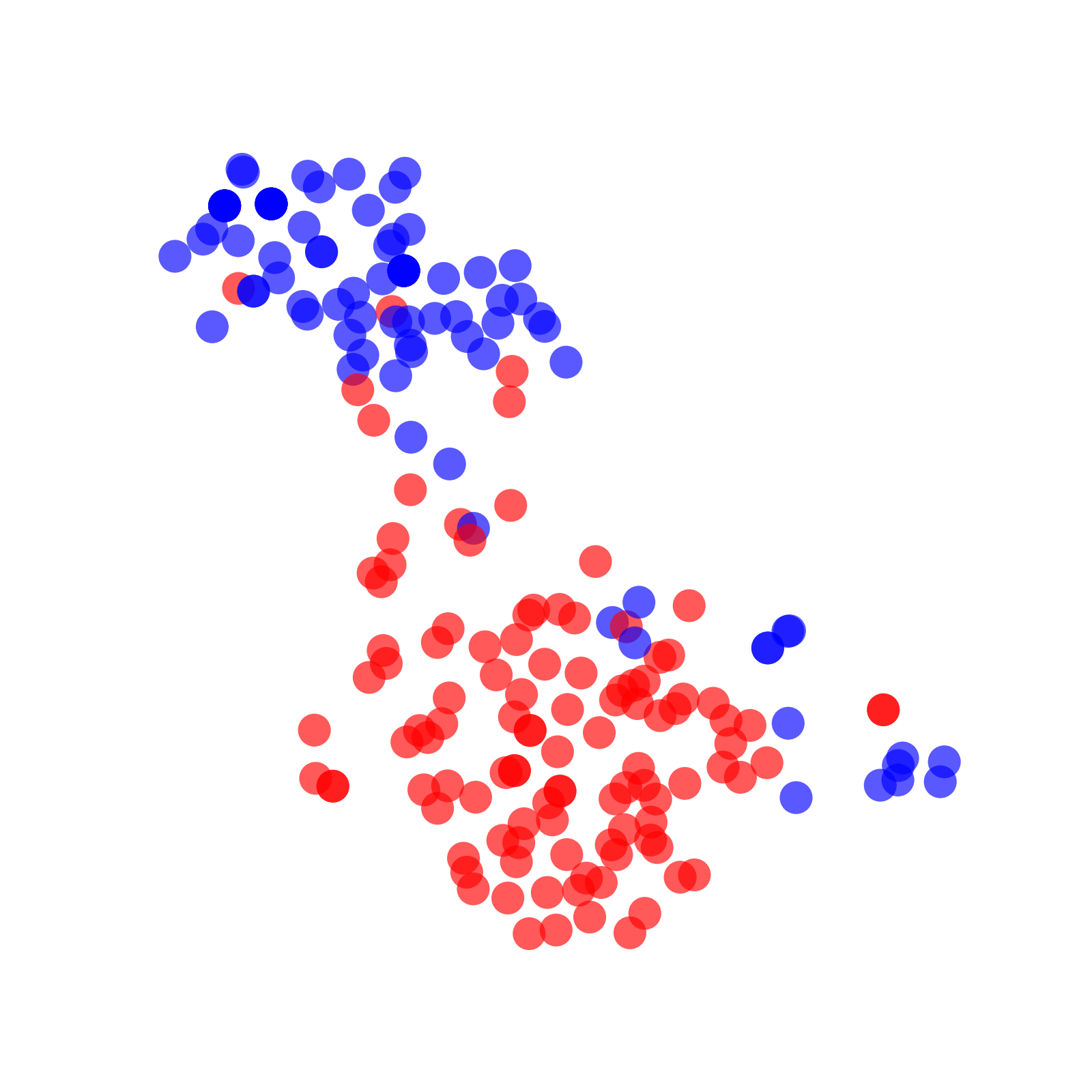}
         \caption{SPOTFAKE}
         \label{FIG:SPOT-cloud}
     \end{subfigure}
     \begin{subfigure}[b]{0.24\textwidth}
         \centering
         \includegraphics[width=\textwidth]{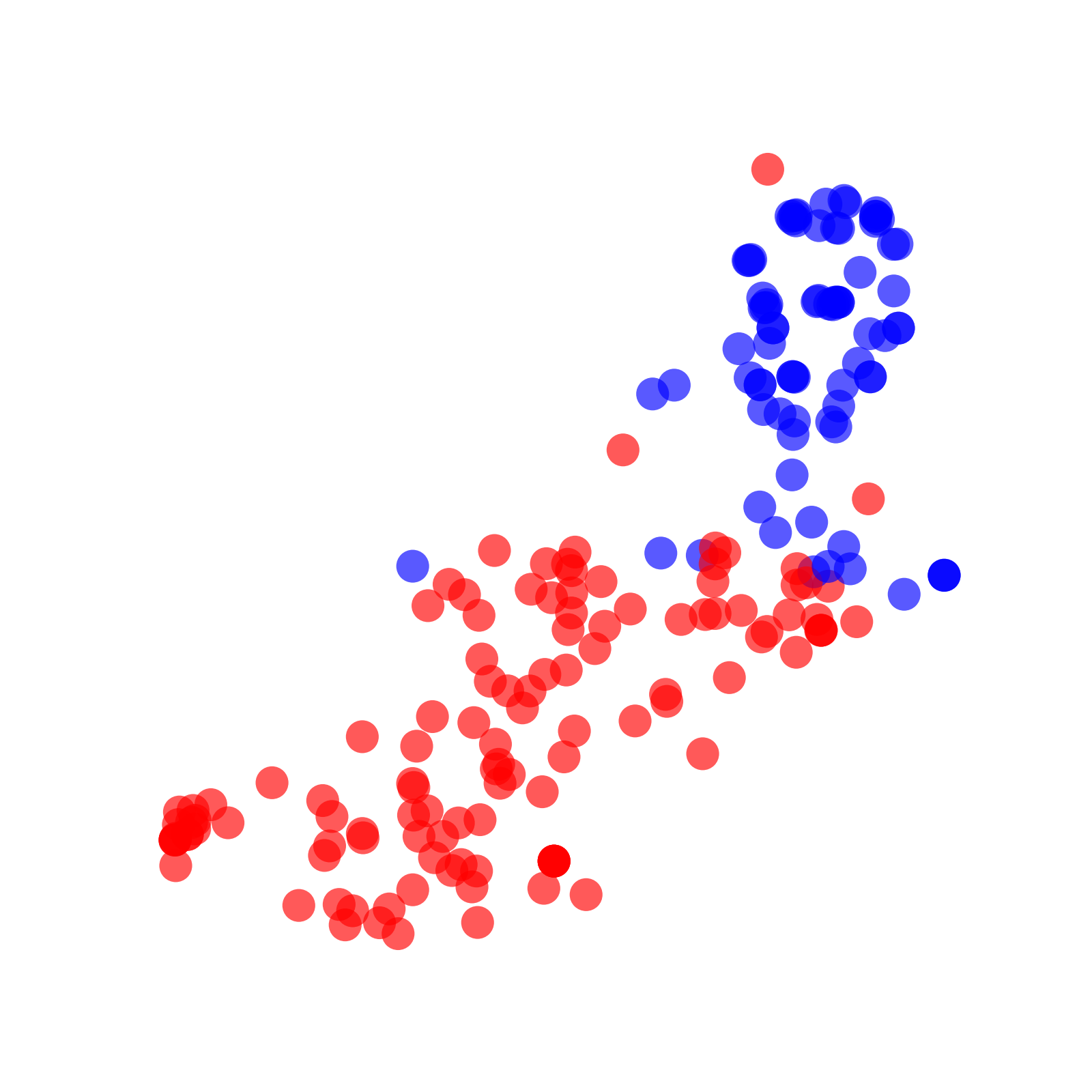}
         \caption{T-BERT}
         \label{FIG:T-cloud}
     \end{subfigure}
     \begin{subfigure}[b]{0.24\textwidth}
         \centering
         \includegraphics[width=\textwidth]{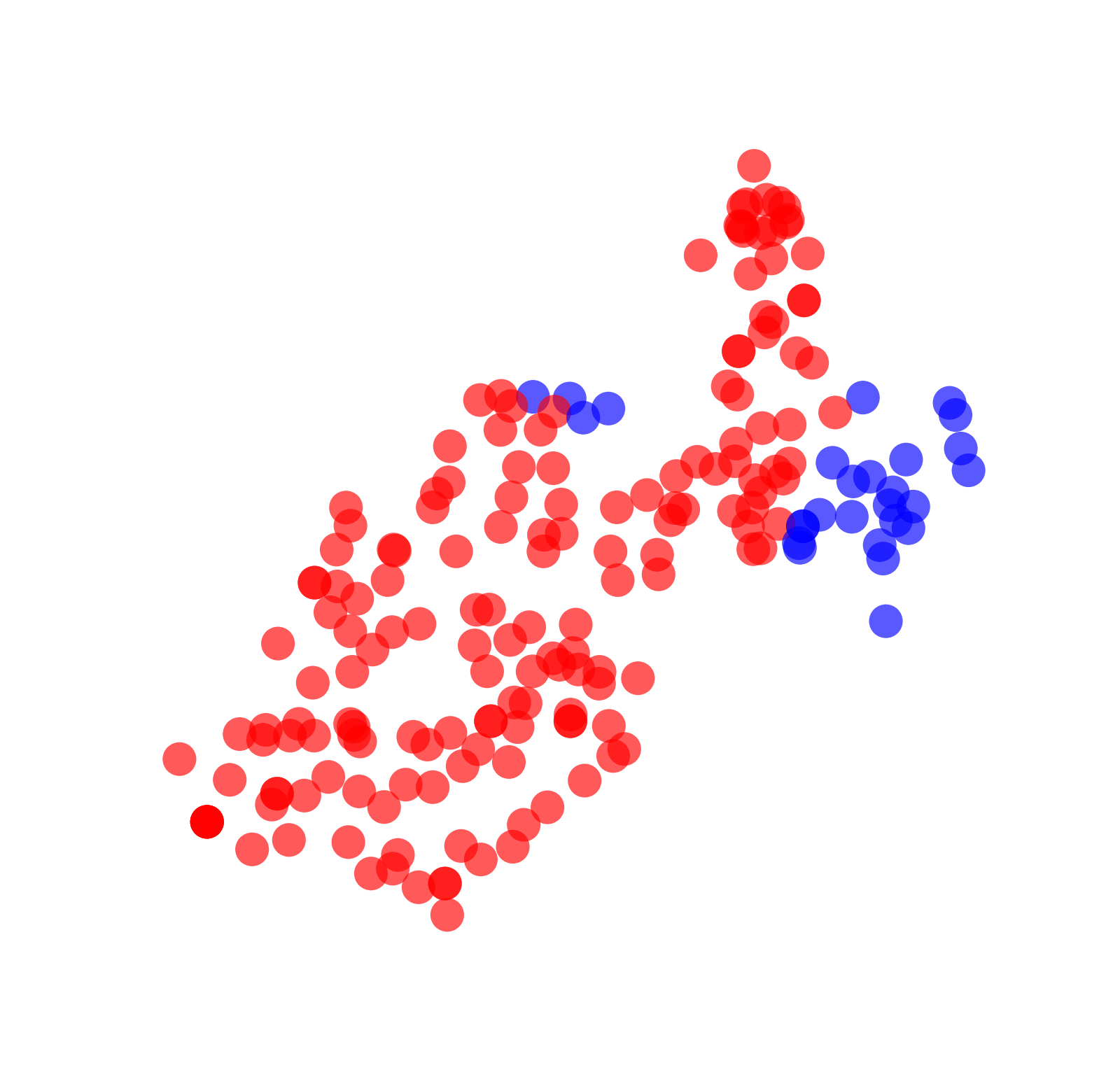}
         \caption{LDA-HAN}
         \label{FIG:LDA-cloud}
     \end{subfigure}
        \caption{T-SNE visualizations of features learned before classifier from M-SAMPLE, D-SAMPLE, C-SAMPLE, RoBERTa, CAFE, SPOTFAKE, T-BERT and LDA-HAN on the test set of PolitiFact in the 2-shot setting.}
        \label{FIG:cloud_all}
\end{figure}

\subsubsection{T-SNE visualization}
We analyze the learned features of the classifiers on the test set of PolitiFact in the 2-shot setting as shown in Figure \ref{FIG:cloud_all}. The reduced dimension learned feature representations of fake and real news are represented by red and blue dots. We observe that the boundary of M-SAMPLE is more sharply defined than that of D-SAMPLE and C-SAMPLE as seen from Figure \ref{FIG:m-cloud}, Figure \ref{FIG:d-cloud} and Figure \ref{FIG:c-cloud}. This suggests that the learned feature representations are more discriminative. Although FT-RoBERTa produces comparable performances in F1 and accuracy, it exhibits some clear misclassified instances in the 2-shot setting. Additionally, the learned feature representations are sparser compared to SAMPLE as shown in Figure \ref{FIG:roberta-cloud}. This implies that in the few-shot scenario, the combination of multimodal features and prompt learning approach outperforms the standard fine-tuning method. We also visualize the feature representations from CAFE and SPOTFAKE, as demonstrated in Figure \ref{FIG:CAFE-cloud} and Figure \ref{FIG:SPOT-cloud}. It is observed that the numbers of misclassified instances are significantly higher than that in prompt learning and fine-tuning methods. Moreover, we find that unimodal methods such as T-BERT and LDA-HAN have the highest number of misclassified instances, suggesting that the multimodal feature can capture more expressive information than the text feature alone, as shown in Figure \ref{FIG:T-cloud} and Figure \ref{FIG:LDA-cloud}.

\section{Discussion}
The proposed SAMPLE framework integrates multiple prompt learning templates with a soft verbalizer to enable the automatic detection of fake news in few-shot and data-rich settings. This section initially analyses the relationships between our approach and existing studies. Additionally, we detail how our proposed SAMPLE approach can positively impact the FND and provide support for real-world applications. Finally, this paper discusses the limitations and future work.		

\subsection{Connections and comparison with previous works}
SAMPLE demonstrates satisfactory performance in detecting fake news in both few-shot and data-rich scenarios. To compare with other approaches in the Fake News Detection (FND) field, traditional approaches fall into one of three categories: (1) unimodal approaches based solely on text or image features \citep{ajao2019sentiment, cao2020exploring, chen2021image}; (2) multimodal approaches that assimilate textual and visual features via either pre-trained models or deep learning representation \citep{kim2014convolutional, tuan2021multimodal}; and (3) standard fine-tuning approaches that fine-tune pre-trained unimodality models with task-specific data \citep{devlin2018bert, nguyen2020bertweet}.

In this study, SAMPLE encompasses a hybrid of approaches (2) and (3), although different from the standard fine-tuning method due to its prompt learning algorithm. Fine-tuning may achieve optimal performance, but consumes a significant amount of memory as it updates the entire set of model parameters for a task-specific objective. Conversely, prompt learning, which leverages a natural language prompt to query a language model, maintains similarity with pre-training and shows comparable performance, particularly with limited training instances. By contrasting the results of the standard fine-tuning with those of SAMPLE, the experimental findings confirm the aforementioned reasoning, as depicted in Table \ref{tab:2}.

Prior multimodal approaches, such as CAFE and SAFE, relied on external cross-modal modules to align and measure disparate unimodality features. Nevertheless, such external modules require an adequate quantity of training instances to capture cross-modal correlations and lead to inadequate performance in the few-shot setting. Consequently, our new proposal offers a similarity-aware multimodal feature fusion methodology that exploits CLIP's pre-training strategy. CLIP utilizes numerous image-text pairs to learn the integration of multimodal semantics. Moreover, the standardization of cross-modal feature correlations incorporates a Sigmoid function to determine the semantic similarity between text and image inputs. An ablation study investigated our approach in the few-shot setting, as depicted in Table \ref{tab:4}, and the resulting data indicates a significant enhancement in few-shot performance due to the combination of prompt learning and the proposed similarity-aware multimodal fusion process.

\subsection{Contributions to future research}
We introduce a novel FND framework, SAMPLE, for identifying fake news using prompt learning. Although prompt learning has demonstrated high performance in numerous classification tasks, integrating different prompting strategies with multimodal features remains underexplored. This paper presents a promising method that achieves strong results and can serve as a significant baseline for future multimodal FND research.

In contrast, traditional multimodal FND systems typically necessitate large amounts of training data to attain satisfactory performance levels. However, obtaining annotated data is challenging in real-world settings. This paper demonstrates that SAMPLE offers comparable results, particularly in few-shot scenarios, indicating its capability to detect fake news in real-world situations. Moreover, the approach that fuses similarity-aware multimodal features with prompt learning holds potential for future similar classification tasks.

\subsection{Limitations and future work}
The present study has several limitations. Firstly, SAMPLE solely focuses on investigating the effects of the soft verbalizer that is designed to identify appropriate label words from the vocabulary automatically. Nevertheless, optimizing the soft verbalizer in a broader vocabulary under low-data conditions remains a considerable challenge, indicating that further adaptive modifications are required to enhance the overall performance. Secondly, the newly proposed multimodal fusing method is based on a similarity-aware strategy that aims to reduce noise injection in less correlated cross-modal features. Nonetheless, it does not explicitly address the uncorrelated cross-modal relations. Thirdly, there is still a need for discovering more multimodal FND approaches to other modalities like news entities and social networks.

Several studies have indicated that the selection of verbalizers considerably affects performance. Notably, manual verbalizers \citep{schick2021s} rely on task-specific prior knowledge and intensive labour work to identify label words representing classes. On the other hand, while the soft verbalizer \citep{shin2020autoprompt, gao2021making} attempts to ease this process, it remains challenging to optimize it adequately for a large vocabulary in low-data settings. Moreover, the knowledgeable prompt-tuning approach \citep{hu2022knowledgeable} utilizes external knowledge bases to expand the coverage of the label words and reduce the bias associated with manual verbalizers. Investigating the impact of verbalizers will be part of our future work. Additionally, integrating other modalities such as news entities, topics and social networks can extend the multimodal fusing method in the future.

\section{Conclusion}
This paper presents a novel similarity-aware multimodal FND framework named SAMPLE that utilizes prompt learning. To mitigate the data insufficient issue, SAMPLE incorporates three popular prompt templates: discrete prompting, continuous prompting and mixed prompting to the original input text, and employs the pre-trained language model RoBERTa to acquire text features from the prompt. Furthermore, the pre-trained CLIP model is used to obtain the input texts, input images, and their semantic similarities. To address semantic gaps and improve the collaboration between image and text modalities, we introduce a similarity-aware multimodal feature fusing approach that applies standardization and a Sigmoid function to adjust the intensity of the final cross-modal representation. Finally, by feeding the multimodal feature into a fully-connected layer, we project the feature to obtain the word distribution mapped to the specific news class.

We evaluate the proposed approach by conducting a multimodal FND experiment on two benchmark datasets, and extensively compare SAMPLE's performance with unimodal, multimodal, and standard fine-tuning approaches. Our experimental results demonstrate that SAMPLE's performance is superior to previous methods, regardless of the few-shot or data-rich settings. Moreover, our results show that, although image modality provides meaningful information, the uncorrelated cross-modal features might harm the FND performances especially when the training instances are small. Additionally, each component of our approach, particularly the standardized multimodal feature fusing module, helps unimodal features from pre-trained models collaborate more effectively in mining crucial features for FND.

\printcredits


\bibliographystyle{cas-model2-names}

\bibliography{cas-sc-template.bib}

\newpage

\appendix

\textbf{\Large{Appendix}}

\section{Prompt engineering for discrete templates}
\label{app:1}

In order to assess the impact of various templates on performance, we have created discrete templates, which are illustrated in Table \ref{tab:5}. Due to the time and cost involved in prompt engineering, we have restricted our study to only five discrete templates in this paper. Subsequently, we choose the discrete template that attains the highest F1 score as the final template in our experiment.

\begin{table}[width=.7\linewidth,cols=4,pos=h]
\caption{The prompt engineering for the discrete templates. All experiments are conducted on the Politifact with fixed seed in 2-shot and alpha=0.8 settings.}\label{tab:5}
\begin{tabular*}{\tblwidth}{@{} LLL @{} }
\toprule
Prompt & F1 & Acc \\
\midrule
This is $<mask>$. & 0.41 & 0.43 \\
This is $<mask>$ news. & 0.41 & 0.47 \\
This news is $<mask>$.  & 0.39 & 0.44 \\
This is a piece of $<mask>$ news.  & 0.43 & 0.45 \\
This is a piece of news with $<mask>$ information. & 0.46 & 0.51 \\
\bottomrule
\end{tabular*}
\end{table}

\section{Comparing with different initialization for continuous templates}
\label{app:2}


The study compares three initialization methods for the $<soft>$ token in the continuous template as demonstrated in Figure 5. The "Random" initialization method initializes the $<soft>$> tokens randomly. The "FC" method reparameterizes the $<soft>$ tokens with another trainable matrix and forward propagates it through an FC layer \citep{li-liang-2021-prefix}. The "LSTM" method feeds the $<soft>$ token through an LSTM layer and employs the outputs as the trainable vectors \citep{liu2021gpt}. Although the performances of the three initialization methods in terms of F1 and accuracies are slightly affected, the study also noted that the "FC" and "LSTM" initializations result in later convergence of validation loss than the "Random" initialization as they require additional training to obtain the $<soft>$ vectors.

\begin{table}[width=.7\linewidth,cols=4,pos=h]
\caption{Different initialization for soft templates. All experiments are conducted on the Gossipcop with fixed seed in 8-shot and alpha=1 settings.}\label{tab:6}
\begin{tabular*}{\tblwidth}{@{} LLL @{} }
\toprule
Init methods & F1 & Acc   \\
\midrule
Random & 0.47 & 0.55  \\
FC & 0.47 & 0.51 \\
LSTM & 0.45 & 0.49 \\
\bottomrule
\end{tabular*}
\end{table}

\section{Comparison between the CLIP and the pre-trained unimodal models for feature extraction}
\label{app:3}

We evaluated the semantic similarity between text and image features from various pre-trained models. We use the BERT model and VGG-19 to extract features from each training sample and then calculate the average to determine semantic similarity for real and fake news. Similarly, we employ the CLIP text transformer and vision transformer to extract unimodal features and calculate their semantic similarity. We also increase the number of samples to observe any changes in semantic similarity. Finally, we scale the values on the axes logarithmically to represent differences since unimodal model differences are small. 

Our experimental findings indicate that the text and image features extract from the CLIP model are more consistent than those from unimodal models, as shown in Figure \ref{app:3}. This can be attributed to the capacity of CLIP to learn multimodal representations through joint training and its utilization of a contrastive loss function to distinguish relevant pairs from irrelevant ones. As a result, the semantic similarity of real news (CLIP\_TRUE) was consistently higher than that of fake news (CLIP\_FAKE) regardless of the number of samples. In contrast, the BERT-VGG19 combination separately extracts features from text and images, which could lead to more noise in feature extraction.

\begin{figure}[h]
    \centering
	\includegraphics[scale=.6]{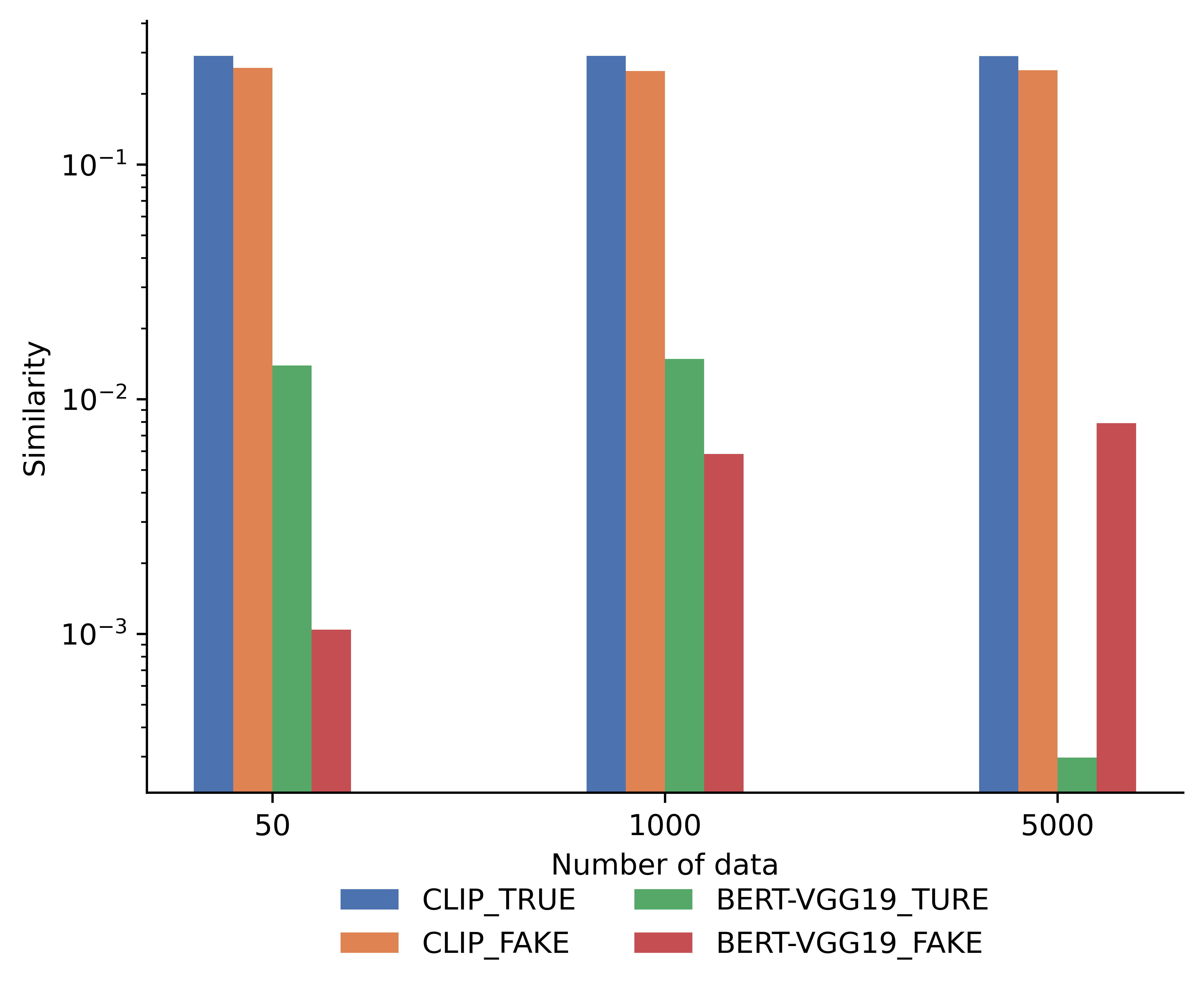}
	\caption{Logarithmically scaled semantic similarity comparison between the pre-trained models.}
	\label{FIG:app3}
\end{figure}





\end{document}